\documentclass[11pt]{article}

\usepackage[letterpaper,margin=1in,headsep=0.25in]{geometry}
\usepackage[utf8]{inputenc}
\usepackage[T1]{fontenc}
\usepackage{lmodern}
\usepackage{microtype}
\usepackage{amsmath,amssymb}
\usepackage{booktabs}
\usepackage{multirow}
\usepackage{graphicx}
\usepackage{xcolor}
\usepackage{url}
\usepackage{hyperref}
\hypersetup{
  colorlinks=true,
  linkcolor=blue!50!black,
  citecolor=blue!50!black,
  urlcolor=blue!50!black,
  pdftitle={Single-Round Vector RAG vs an LLM-Compiled Wiki: A Preregistered Comparison on a Small Multi-Domain Research Corpus},
  pdfauthor={Theodore O. Cochran},
  pdfsubject={Retrieval-augmented generation, LLM-compiled wiki, grounded synthesis},
  pdfkeywords={RAG, LLM wiki, retrieval augmented generation, citation faithfulness, LLM-as-judge, preregistration}
}
\usepackage[capitalise]{cleveref}
\usepackage{natbib}
\newcommand{\rag}{Vector RAG\xspace}
\newcommand{\wiki}{LLM Wiki\xspace}
\newcommand{\opus}{Claude Opus 4.7\xspace}
\newcommand{\gpt}{GPT-5.4\xspace}
\newcommand{\gem}{Gemini 2.5 Pro\xspace}
\newcommand{\Hone}{H\textsubscript{1}\xspace}
\newcommand{\Htwo}{H\textsubscript{2}\xspace}
\newcommand{\Hthree}{H\textsubscript{3}\xspace}
\newcommand{\Hthreea}{H\textsubscript{3}a\xspace}
\newcommand{\Hthreeb}{H\textsubscript{3}b\xspace}
\usepackage{xspace}

\makeatletter
\def\verbatim@font{\small\ttfamily}
\makeatother

\newcommand{\osflink}{\url{https://doi.org/10.17605/OSF.IO/ZEMHP}}

\newcommand{\artifactlink}{\url{https://osf.io/j37b8/}}
\newcommand{\coderelease}{\url{https://github.com/ai4altruism/rag2compare}}

\title{Single-Round Vector RAG vs an LLM-Compiled Wiki: \\
  A Preregistered Comparison on a Small Multi-Domain Research Corpus}

\author{%
  Theodore O. Cochran \\
  AI for Altruism (A4A) \\
  \texttt{theo@ai4altruism.org}
}

\date{}

\begin{document}
\maketitle

\begin{abstract}
We preregistered a comparison of two ways to help an LLM answer questions over a small research corpus: a single-round \rag system and an LLM-compiled markdown wiki browsed by a tool-using agent. Both systems answered the same 13 questions over 24 papers using the same answer-generating model, and their answers were scored by two blinded LLM judges.

The three preregistered predictions, reported here in their registered order, came out one weakly supported, one supported, and one refuted. First, the wiki was predicted to synthesize better across papers, on both cross-paper connection and answer organization. It scored much better at connecting findings, but its organization advantage fell below the registered threshold once both judges' scores were combined, leaving this prediction only weakly supported. Second, RAG was predicted to hold its own on single-fact lookup questions, and it met the registered test on the scored groundedness criterion, though that result was judge-sensitive: the second judge alone would have refuted it, and the preregistered Bayesian check did not corroborate it. Third, the wiki was predicted to be expensive to build and cheap to query, paying for itself after enough questions. The build side held, by roughly two orders of magnitude, but the query side reversed: the wiki spent about 21 times more LLM tokens per query than RAG, so the cheaper queries never arrive and no break-even point exists. That prediction is refuted.

Two exploratory analyses bear on why such comparisons disagree. Against the single-round baseline, the compiled wiki scored much higher on multi-paper synthesis, but a decomposition-retrieval variant of RAG reduced most of that observed gap at lower token cost, so the gap is sensitive to how retrieval is done. The two judges also agree very unevenly: their rank agreement is near zero on holistic groundedness ($\rho = 0.04$), against $\rho = 0.81$ on the most concretely defined criterion. A post-hoc claim-level analysis of citation support was checked against two human annotators on a sample of 100 claims. Its automated scorer agreed with them on 50 to 54\% of claims as first run and on 65 to 69\% once a truncation error was corrected, short of the agreement rule fixed in advance against each annotator. On the annotators' labels, cited claims were supported 78\% of the time for RAG and 78\% or 86\% for the wiki, so the analysis did not establish a citation-support advantage for the wiki.

The conclusion is narrower than a winner. On one model and one small corpus, the compiled wiki organized multi-paper evidence better than single-round retrieval on LLM-judged criteria, at about 21 times the per-query token use, and a decomposition-retrieval baseline recovered most of that advantage at lower cost. Which system appears to win depends on the retrieval baseline, the scoring method, and the judge. Evaluations of this kind should report synthesis structure, citation support, and cost separately, and should check any automated grounding scorer against human labels before comparing systems with it.
\end{abstract}

\section{Introduction}
\label{sec:intro}

An ``LLM Wiki'' turns a research corpus into a cross-linked wiki at ingest time, and at question time the model browses that wiki rather than retrieving raw chunks from the original papers. This architecture was popularized in informal commentary, notably Andrej Karpathy's ``agentic markdown wiki'' framing. The idea suggests three possible tradeoffs: the wiki may help with multi-paper synthesis; rewriting papers into wiki pages may lose source fidelity; and the wiki may move cost from query time to ingest time, with a break-even point after enough questions. Compiled-knowledge architectures have since been compared against retrieval baselines by several groups, with materially different results depending on how the compiled knowledge is served and what is measured (\cref{sec:related}). We do not claim priority; we claim a design, and the contribution of this paper is what a preregistered, blinded, two-judge protocol, with a claim-level citation analysis checked against human annotators, shows about \emph{why} such comparisons disagree.

This paper reports a preregistered, blinded, two-judge comparison of \rag and \wiki on a fixed 24-paper corpus answering 13 evaluation questions in six task categories, which we call tiers. One model, \opus, performs every generating step: it answers the questions for both systems (at \texttt{xhigh}), compiles the wiki, enriches the RAG index at ingest, and atomizes answers into claims for the claim-level analysis. The conclusions are therefore about these implementations on this model, and the wiki's results depend on its compile model as well as its query model. The cross-family judge is \gpt, with \gem for inter-rater reliability; agreement between the two measures consistency between models, not agreement with people. Corpus, question set, rubric, decision rules, and Bayesian model were locked at the OSF preregistration tag before any judge run. The two systems differ in corpus representation, retrieval strategy, and test-time compute at once, so the comparison cannot attribute a difference to the compiled representation alone. Preregistration constrains the analysis but supplies neither statistical power nor representativeness, and every conclusion here is limited to this corpus, question set, configuration, and model.

\paragraph{Contributions.}
\begin{itemize}
\item \textbf{Preregistered comparison.} On LLM-judged scores, the wiki was much better at connecting papers; its advantage in answer organization was weaker after judge adjustment, positive but below the registered \Hone $+2.0$ threshold (the rubric criteria are \texttt{inter\_paper\_mapping} and \texttt{structural\_integrity}). RAG met the preregistered test for single-fact source grounding on \texttt{groundedness} (small-effect-fragile under IRR adjustment). The expected wiki cost advantage failed in the opposite direction on measured token use: the wiki spent about $21\times$ more per query than RAG.
\item \textbf{Evaluation-method finding.} Inter-judge agreement varied sharply across rubric criteria, and the spread is extreme: it was highest on the most operationally concrete criterion and lowest on the most holistic one, Spearman $\rho = 0.81$ (ICC $0.81$) on \texttt{inter\_paper\_mapping}, the one criterion with a countable definition, against $\rho = 0.04$ (ICC $0.04$) on holistic \texttt{groundedness}, where \gem saturates near 10/10 and \gpt uses more of the scale. This suggests criterion design, not judge identity alone, can materially affect LLM-judge reliability; agreement should be measured per criterion rather than assumed uniform across a rubric.
\item \textbf{Exploratory RAG ablation (two-judge).} A decomposition-retrieval variant of RAG reduced most of the observed score gap on cross-paper synthesis, closing $\sim 94\%$ of it on the \Hone subset under the same IRR adjustment the confirmatory analysis uses, or $\sim 88\%$ scored by the primary judge alone, and leaving a remaining wiki advantage smaller than what the registered $+2.0$ threshold would have required. On the corrected claim-level scorer, its cited claims were supported less often than either other arm's.
\item \textbf{Human check of the claim-level scorer, and a correction.} Two annotators scored a stratified sample of 100 cited claims under a decision rule fixed in advance. They agreed with each other on 94. The post-hoc claim-level scorer agreed with them on 50 to 54 as first run and on 65 to 69 once corrected, and met the rule against neither annotator either time; agreement with an adjudicated consensus was not assessed. The check exposed an error in the scoring as first run, which truncated RAG's cited evidence to its first 1{,}500 characters. With the error corrected, and on the annotators' labels, the analysis does not establish that the wiki's cited claims are better supported than single-round RAG's (\cref{sec:results-claim}, \cref{app:correction}).
\end{itemize}

\paragraph{Deviations from the registered procedure.} One concerns timing. The registration states that the RAG run, the wiki-side run, and the judge would not be invoked before registration acceptance. The RAG run and the judge complied; the \textbf{wiki-side run was invoked 77 minutes before the registration timestamp}. We report this here rather than leave it to be found: \cref{sec:methods} gives the full chronology, what it does and does not put at risk, and the frozen-instrument timestamps that bound it. The other is separate and concerns coverage: secondary judging was extended from the registered three-question subset to all 13 questions (\cref{sec:experiments}).

Five preregistered run artifacts and eight post-hoc artifacts, including the rescoring outputs behind the corrected claim-level figures, are deposited on OSF with SHA-256 hashes; the reproducer scripts ship in the supplementary material.

\section{Related Work}
\label{sec:related}

\paragraph{RAG and multi-hop retrieval.} Retrieval-augmented language modeling was introduced by REALM~\citep{guu2020realm} and Lewis et al.~\citep{lewis2020rag} with dense passage retrieval~\citep{karpukhin2020dpr}. Our \rag is a single-round retrieve-then-generate instantiation with multi-query expansion, hybrid retrieval, Cohere reranking, and CRAG-inspired~\citep{yan2024crag} corrective validation; it does not perform retrieval-during-generation (cf.\ IRCoT~\citep{trivedi2023ircot}, FLARE~\citep{jiang2023flare}). Multi-hop benchmarks (HotpotQA~\citep{yang2018hotpotqa}, MuSiQue~\citep{trivedi2022musique}, MultiHop-RAG~\citep{tang2024multihoprag}) document that single-round retrieval is insufficient for connected reasoning across documents.

\paragraph{Hierarchical and abstractive memory.} The \wiki architecture sits closest to compiled abstractive memory: an offline LLM-built representation queried in lieu of (or alongside) raw chunks. RAPTOR~\citep{sarthi2024raptor} builds a recursive cluster-summary tree; RECOMP~\citep{xu2024recomp} compresses retrieved documents; GraphRAG~\citep{edge2024graphrag} builds an entity graph with community summaries; knowledge-graph variants include REANO~\citep{fang2024reano} and GNN-RAG~\citep{mavromatis2025gnnrag}. Our wiki is conceptually an LLM-built version of this family with a markdown-rendered intermediate representation. OpenScholar~\citep{asai2026openscholar} and PaperQA2~\citep{skarlinski2024paperqa2} apply similar primitives at much greater scale.

\paragraph{Contemporaneous compiled-knowledge comparisons, and why they disagree.} Three studies bear directly on the present comparison, and their results do not line up with each other or with ours. \citet{huerta2026wicer} compiles a wiki memory and serves it as full context via KV caching over 17 RepLiQA domains and 6{,}800 questions, reporting that full-context inference beats RAG on curated knowledge (4.38 vs 4.08 of 5) but degrades below RAG at larger scale through attention dilution, and that blind compilation loses substantial quality before iterative refinement recovers it. \citet{zhang2026policywiki} compares an offline-compiled wiki against BM25 RAG on 28 cross-institutional policy queries with a shared generator, and reports the wiki ahead on citation recall (0.242 vs 0.095) while consuming a \emph{fraction} of the retrieval baseline's query-time context. \citet{chan2024cag} argues more broadly that for knowledge bases small enough to fit in context, preloading and caching removes the retrieval step altogether. Our per-query result runs opposite to \citeauthor{zhang2026policywiki}'s. The results are not directly contradictory, because the studies evaluate substantially different operating points, differing in corpus, task, generator, retrieval baseline, scale, and metrics as well as in architecture. Serving strategy is an especially salient difference: their wiki is \emph{served} as retrieved pages, and \citeauthor{chan2024cag}'s and \citeauthor{huerta2026wicer}'s from a cached context window, whereas ours is browsed by a tool-using agent that pays for its own traversal, up to 30 turns of it. We cannot isolate that factor across studies. This is the paper's central point in miniature. ``Wiki'' and ``RAG'' name families of systems, not operating points, and a comparison can reverse its cost verdict under a different serving strategy. It is also why we report synthesis structure, claim-citation alignment, and cost as three separate axes rather than one.

\paragraph{Citation-faithfulness and LLM-as-judge.} ALCE~\citep{gao2023alce} introduced citation-precision/recall benchmarks for long-form answers; \citet{liu2023verifiability} found commercial generative search systems often produce fluent but unsupported answers; our claim-level grounding analysis follows this protocol family. \citet{zheng2023mtbench} document position, verbosity, and self-enhancement biases in LLM judges; G-Eval~\citep{liu2023geval} reports improved human alignment under structured form-filling but notes possible bias toward LLM-generated text; \citet{wang2024fairevaluators} show ranking can change with response order. Our IRR ceiling-judge finding (\cref{sec:results-irr}) is consistent with this literature.

\section{Methods}
\label{sec:methods}

\paragraph{Corpus and questions.} 24 peer-reviewed papers across three domains (8 each: AI ethics \& law, climate science, precision medicine), 2017--2026. 13 evaluation questions across six tiers: chronological, conflict, multi-hop, emergence, and policy ($n=2$ each), and bias-check ($n=3$, with extra power for \Htwo). Each question carries an expected-bias label in the question file frozen at the preregistration tag: the conflict, multi-hop, emergence, and policy questions are labeled wiki-favoring, the two chronological questions neutral, and the three bias-check questions RAG-favoring. \Cref{app:decomp-tiers} describes how the tiers were constructed. Inclusion: peer-reviewed or arXiv-preprint research papers; exclusion: review-only summaries, books, non-research artifacts. Corpus list with DOIs and full question text in the OSF deposit.

\paragraph{\rag.} RAG retrieves passages once, reranks and validates them, then writes the answer. Implementation: header-aware chunking (target 512 tokens, 50-token overlap), per-chunk contextual enrichment, multi-query expansion, dense+sparse hybrid retrieval, Cohere reranking to top-5, and CRAG-inspired~\citep{yan2024crag} corrective validation. Exact model versions, prompts, and parameters for every stage are in the supplementary material. \opus performed the contextual enrichment; the deposited ingest artifact does not record the enrichment model, and the run overrode a different default in the code. ``Single-round'' means no retrieval-during-generation (cf.\ \citep{trivedi2023ircot, jiang2023flare}); the pipeline is otherwise multi-stage.

\paragraph{\wiki.} The wiki system first compiles the papers offline, with \opus, into a persistent cross-linked markdown wiki (entities, concepts, sources, and analyses with cross-references). At question time, a tool-using agent lists pages, reads selected pages, and submits an answer via three tools (\texttt{list\_pages}, \texttt{read\_page}, \texttt{submit\_answer}; \texttt{MAX\_TURNS}=30, no query-side prompt caching). Two minor telemetry caveats ($\le 7\%$ thinking-token over-count; $\le 6.5\%$ upper-bound caching saving forgone on the query side) do not materially affect the per-query comparison.

Both systems use the same query model (\opus at \texttt{xhigh}), corpus, and evaluation set; the comparison is fully crossed and within-question paired. This is a comparison of two practical architectures, not a clean causal isolation of ``knowledge organization'': the systems differ along multiple axes (offline LLM rewriting; compiled markdown vs vector store; tool-loop vs single-round retrieval; multi-call vs single-call context). Corpus representation, retrieval strategy, and test-time compute therefore change together, and no difference reported below can be attributed to the compiled representation alone.

\paragraph{Judging.} Primary: \gpt at \texttt{medium} reasoning, cross-family from the answer model. IRR: \gem, same prompt, fixed seed offset. Per-question \{RAG, Wiki\}~$\to$~\{A, B\} blinding via \texttt{random.Random(seed=42)} (primary) / \texttt{(seed=43)} (secondary). The judge prompt instructs the model to score each system independently on each criterion. Rubric: four 1--10 anchored criteria with explicit anchors at 1/5/10: \texttt{groundedness} (claims trace to source material), \texttt{structural\_integrity} (unified narrative vs concatenated fragments), \texttt{conflict\_awareness} (naming of contradictory findings), \texttt{inter\_paper\_mapping} (multi-hop synthesis across $\ge 2$ papers). Agreement between the two judges measures consistency between models, not agreement with people. Only the post-hoc claim-level scorer has been checked against human annotators (\cref{app:correction}); the rubric results are comparisons of LLM-judged scores.

\paragraph{Preregistration and decision rules (paraphrased).}
\begin{itemize}
\item \Hone (synthesis): Wiki $-$ RAG $\ge +2.0$ on \texttt{inter\_paper\_mapping} \emph{and} \texttt{structural\_integrity} for the multi-hop+emergence tier ($n=4$). Supported if both clear; weakly supported if both positive but at least one below; refuted if either negative.
\item \Htwo (point-source): RAG $-$ Wiki $\ge 0$ on \texttt{groundedness} for the bias-check tier ($n=3$).
\item \Hthree (cost): both $T_\text{ingest}[\text{wiki}] > T_\text{ingest}[\text{rag}]$ AND $T_\text{query}[\text{rag}] > T_\text{query}[\text{wiki}]$.
\end{itemize}
The full preregistration is available at \osflink. The preregistration is explicit that the primary analyses are directional comparisons of means against magnitude thresholds, not frequentist tests, because $n$ is small and LLM scores have non-i.i.d.\ error structure. Two robustness checks are also preregistered (bootstrap percentile CI; Bayesian model with $\mu \sim \mathcal{N}(0,4)$, $\sigma \sim \text{HalfNormal}(4)$, NUTS).

\paragraph{Chronology, and one deviation from the registered starting rule.} Because the tier design labels four question tiers wiki-favoring, one neutral, and one RAG-favoring, the order of events matters to how much the registration constrains. All times UTC on 2026-05-06 unless noted.

\begin{table}[h]
\caption{Execution chronology. Times are the artifacts' own \texttt{started\_at} fields, which each runner captures at invocation, and git author timestamps for the commits. Every row is checkable against the deposited artifacts; the supplementary material gives the commit identifiers.}
\label{tab:chronology}
\centering
\small
\begin{tabular}{lll}
\toprule
Time (UTC) & Event & Evidence \\
\midrule
05-05 01:41 & Questions and rubric frozen & last commit to \texttt{questions/rubric.yaml} \\
20:08 & Design freeze & the commit carrying the prereg tag \\
20:29 & RAG-side ingest invoked & \texttt{ingest-...T202904Z} \\
20:55 & \textbf{Wiki-side run invoked} & \texttt{wiki-run-...T205500Z} \\
\textbf{22:12} & \textbf{OSF registration} & \texttt{date\_registered}, reg.\ \texttt{zemhp} \\
22:16 & RAG-side run invoked & \texttt{run-...T221602Z} \\
22:56 & Judge invoked & \texttt{judged-...T225645Z} \\
\bottomrule
\end{tabular}
\end{table}

The registration's starting rule states that ``the 13-question RAG run, the Wiki-side run, and the judge will not be invoked before registration acceptance,'' and separately discloses that RAG-side ingestion was already running. \textbf{The wiki-side run does not satisfy that rule}: it was invoked 77 minutes before the registration timestamp and completed 13 minutes after it. The RAG run and the judge do satisfy it, at $+3$ and $+44$ minutes. We report this rather than leave it to be discovered, and we have not restated the rule to fit what happened.

The deviation weakens the procedural preregistration claim, and we state the weakening rather than argue around it: the wiki run had begun before the OSF registration became immutable, and incidental exposure to its output cannot be excluded. Wiki answer text is outcome information even though no judge score existed yet.

Three facts bound the scope for outcome-contingent adaptation, and each is externally checkable. The question set and rubric were committed 43 hours earlier and unmodified after. The registered decision rules and thresholds are in the same frozen commit, which the registration cites by tag. And no judge score of any kind existed until 44 minutes \emph{after} registration, so the registered analysis plan was fixed before any score was observable. These timestamps substantially constrain, but do not eliminate, the scope for the design to have followed the data.

\paragraph{IRR-triggered adjustment.} If the maximum primary--secondary score delta exceeds 2 on any criterion, that criterion's \texttt{mean\_tier} values are recomputed using the mean of the two judges' per-question scores in place of the primary-judge-only mean for the \Hone/\Htwo tests.

\section{Experimental setup}
\label{sec:experiments}

\paragraph{Artifacts.} Five preregistered run artifacts plus eight post-hoc artifacts (claim-level grounding on the original run; the decomp ablation run, its judge run, and its grounding; the two-judge rejudge of the ablation; and three claim-level rescoring outputs: the main arms and the decomposition arm on full passages, and a capped rerun control) are deposited on OSF at \artifactlink{} with SHA-256 hashes. Every reproducer script cited in this paper, including the analysis that regenerates the reported values and the agreement statistics in \cref{tab:irr-agreement}, ships in the supplementary material accompanying this submission, which is self-contained and requires no repository access. \emph{Manipulation check.} All 26 (question, system) cells used \texttt{claude-opus-4-7} at \texttt{xhigh}; zero exclusions, zero ingest/query/judge failures.

\paragraph{Cost-accounting scope.} \Hthree follows the preregistered token-accounting (sum of prompt + completion at ingest; prompt + completion + thinking at query) and excludes embedding spend, reranker spend, vector-store storage, and dollar normalization. \emph{Terminology.} Because of those exclusions, and because the recovered ingest figure weights the input side for caching while leaving output tokens at face value, the quantity this paper measures is \textbf{LLM-token usage}, a token-accounting proxy for cost, and not billable cost. We write ``cost'' for it only where the token basis is unambiguous from context; ratios such as the $21\times$ per-query figure are ratios of tokens, and they would not survive unchanged into dollars under provider pricing that bills token classes at different rates. \emph{Prompt-cache caveat (load-bearing).} The wiki ingest artifact rolls \mbox{$\text{uncached} + \text{cache\_creation} + \text{cache\_read}$} into one field. Anthropic bills cache reads at $\sim 10\%$ of base input rate, so the gross figure overstates billable cost by an order of magnitude on cache-heavy workloads. The \Hthreeb test is captured uncached on both sides and needs no correction. \Hthreea is adjudicated on the registered token metric directly from the deposited artifacts; because that metric's economic weight is uneven across token classes, \cref{sec:results} additionally reports a post-hoc billing-weighted sensitivity analysis from per-session components that reconcile exactly to the deposited aggregate, as a range over the two possible cache-write rates.

\paragraph{Robustness checks (preregistered).} \emph{Bootstrap}: 95\% percentile CI, 10{,}000 resamples per criterion $\times$ tier. \emph{Bayesian}: per criterion $\times$ tier, $\text{score\_diff} \sim \mathcal{N}(\mu, \sigma^2)$ with $\mu \sim \mathcal{N}(0, 4)$, $\sigma \sim \text{HalfNormal}(4)$, NUTS (4 chains $\times$ 2{,}000 iter $\times$ 2{,}000 warmup, \texttt{random\_seed=42}); all fits met the preregistered convergence criteria ($\hat R \le 1.003$, min ESS $\ge 1{,}907$). The Bayesian ``strongly corroborated'' threshold is $P(\mu \ge \theta) \ge 0.95$ ($\theta = +2$ for \Hone, $\theta = 0$ for \Htwo).

\paragraph{IRR adjustment.} The preregistration specified a 3-question IRR subset; during execution we extended secondary judging to all 13 questions, a disclosed rigor-increasing deviation. The max-delta trigger fires on all four criteria under both the prereg $n=3$ and the expanded $n=13$ coverage; the adjustment rule is invariant to the deviation.

\paragraph{Post-hoc claim-level grounding.} For each of the 26 answer cells, \opus (medium adaptive thinking) atomizes the answer into atomic claims with \texttt{cited\_source\_idx}; for each cited claim, \gpt (medium, cross-family) scores against the cited chunks as \texttt{supported} / \texttt{partial} / \texttt{contradicted} / \texttt{unsupported}. Uncited claims auto-receive \texttt{unsupported}. Full prompts in \cref{app:claimgrounding}. \textbf{Correction.} As first run, the scoring script cut every cited passage to its first 1{,}500 characters before the scorer saw it. RAG cites chunks of about 2{,}500 characters and the wiki cites page excerpts of about 430, so the cut reached 128 of the 132 cited single-round RAG claims, 200 of the 219 cited decomp-RAG claims, and none of the 241 cited wiki claims. Every claim-level figure in this version comes from rescoring the same claims on full passages, with the same prompt and scorer model; \cref{app:correction} gives the figures as first reported, a capped rerun that separates the cut from rerun noise, and a human check of the scorer. The atomization step saw each source cut to 600 characters; it takes citations from the answer's own attributions and was not rerun.

\section{Results}
\label{sec:results}

We report each hypothesis under the \textbf{primary-only} reading (the pre-IRR baseline) and the \textbf{IRR-adjusted} reading (judge-mean recomputation; see \cref{sec:methods}). The trigger and adjustment rule were both preregistered. The IRR-adjusted reading is the more conservative of the two and is the basis for our headline verdicts.

\subsection{Confirmatory analyses}

\paragraph{\Hone: Synthesis advantage ($n=4$).} Per-question paired diffs (Wiki $-$ RAG) on the multi-hop + emergence subset are computed by the preregistered analysis script in the supplementary material, and the primary judge's per-question scores are in \cref{app:per-question}; the subset means are \texttt{structural\_integrity} $+2.000$ primary / $+1.250$ secondary / $+1.625$ judge-averaged, and \texttt{inter\_paper\_mapping} $+6.250$ / $+7.000$ / $+6.625$.

Primary-only: $\Delta_{\text{struct}} = +2.000$ (lands exactly at threshold), $\Delta_{\text{mapping}} = +6.250$ (clears). Both criteria $\ge +2.0$ $\Rightarrow$ \emph{supported}. \emph{IRR-adjusted (headline)}: $\Delta_{\text{struct}} = +1.625$ (misses by 0.375), $\Delta_{\text{mapping}} = +6.625$ (clears) $\Rightarrow$ \emph{weakly supported} per the preregistered three-way decision rule. The \texttt{inter\_paper\_mapping} advantage is large and never close to the threshold; the \texttt{structural\_integrity} advantage is threshold-sensitive.

With $n=4$ the most legible robustness check is to drop each question in turn. On \texttt{inter\_paper\_mapping} the judge-averaged mean stays between $+5.833$ and $+7.833$, never approaching $+2.0$, so that verdict does not depend on any single question. On \texttt{structural\_integrity} it moves between $+1.000$ and $+2.000$: dropping \texttt{T3-rwd-validity-for-side-effects} lands exactly on the threshold and would flip the verdict from weakly supported to supported, and dropping \texttt{T4-eelgrass-mrv-gaps} halves the effect. The \Hone verdict is therefore two separate findings with very different evidential weight, and we treat the leave-one-out spread, not the third decimal place of the posterior, as the honest summary of what $n=4$ supports.

\paragraph{\Htwo: Point-source parity ($n=3$).} Per-question paired diffs (RAG $-$ Wiki) on \texttt{groundedness} for the bias-check tier, primary / secondary: B1 $+1$/$+3$, B2 $+3$/$-6$, B3 $+2$/$+1$. Means are $+2.000$ primary, $-0.667$ secondary, $+0.667$ judge-averaged.

The B2 secondary entry is the largest single judge disagreement in the experiment: \gpt scored RAG=9 / wiki=6, \gem scored RAG=4 / wiki=10. Both readings nonetheless return $\Delta \ge 0$, satisfying the preregistered decision rule. \emph{Primary-only}: $\Delta = +2.000$ $\Rightarrow$ \emph{supported}. \emph{IRR-adjusted}: $\Delta = +0.667$ $\Rightarrow$ \emph{supported}.

We report this verdict as registered and do not relabel it after the fact, but the evidence behind it is thin and we state its shape rather than only its outcome. \Htwo \emph{met the preregistered point-estimate criterion, and was judge-fragile and not strongly corroborated by the preregistered Bayesian check}: the judge-averaged effect is $+0.667$ on a 1--10 scale, the secondary judge alone returns $-0.667$ and would refute, the Bayesian posterior gives $P(\mu \ge 0) = 0.660$ against the registered corroboration bar of $0.95$, and the criterion it rests on is the one with near-zero inter-judge rank agreement (\cref{tab:irr-agreement}). Leave-one-question-out over the $n=3$ subset keeps the sign but reaches the threshold exactly: dropping \texttt{B1} gives $\Delta = +0.000$, \texttt{B3} gives $+0.250$, \texttt{B2} gives $+1.750$. \Cref{sec:results-claim} reports a post-hoc analysis that nuances the underlying mechanism.

\paragraph{\Hthree: Cost asymmetry.} The preregistration expected the wiki to be expensive to build but cheap to query. The clean query-side data showed the opposite: the wiki cost more per query, so the amortization story cannot hold. Both sides are adjudicable directly from the registered token accounting. We then report a post-hoc economic sensitivity analysis, because the wiki's registered ingest count is dominated by cache reads that a provider bills at a fraction of the base rate, and we want the verdict to be shown robust to that rather than resting on a count whose economic meaning is uneven.

\begin{table}[h]
\centering
\small
\begin{tabular}{@{}lr@{}}
\toprule
Quantity & Value \\
\midrule
$T_\text{query}[\text{rag}]$, 13 questions (prompt + completion + thinking) & $78{,}093$ \\
$T_\text{query}[\text{wiki}]$, 13 questions (prompt + completion + thinking) & $1{,}651{,}357$ \\
\midrule
$T_\text{ingest}[\text{rag}]$, 24 papers (prompt + completion) & $1{,}401{,}996$ \\
$T_\text{ingest}[\text{wiki}]$, \textbf{registered metric} (\texttt{tokens\_in} + \texttt{tokens\_out}) & $904{,}884{,}660$ \\
\quad\emph{post-hoc} billing-weighted input, 5-min / 1-hour TTL & $135{,}702{,}160$ / $161{,}494{,}660$ \\
\midrule
H3a supported $\bigl(T_\text{ingest}[\text{wiki}] > T_\text{ingest}[\text{rag}]\bigr)$ & \textbf{True} (registered, and both weightings) \\
H3b supported $\bigl(T_\text{query}[\text{rag}] > T_\text{query}[\text{wiki}]\bigr)$ & \textbf{False} ($\mathbf{21\times}$ opposite direction) \\
\bottomrule
\end{tabular}
\end{table}

The wiki query-side harness explicitly disabled prompt caching (companion §4.3), and RAG per-query is similarly uncached, so the per-query figures are directly observed uncached token counts and require no cache-class correction. The $21\times$ per-query gap is a \emph{token} ratio, not a dollar-cost ratio.

\Hthree as a conjunctive hypothesis requires both \Hthreea and \Hthreeb to be supported, so \Hthreeb's opposite-direction result alone is enough to refute it. The wiki cost intuition was that pre-compilation pays a large up-front fee in exchange for cheaper queries, breaking even at some N. The opposite happened: the user pays \emph{more} per query, not less, so the preregistered wiki-amortization mechanism is impossible under the observed per-query costs. The preregistered crossover-queries formula's denominator $(T_\text{query}[\text{rag}] - T_\text{query}[\text{wiki}])/13 = -121{,}020$ is negative; the formula returns a negative N, mathematically encoding ``no positive-query crossover''.

\paragraph{\Hthreea{} on the registered metric.} The preregistered accounting sums prompt and completion tokens, and on that metric \Hthreea is adjudicated by the deposited artifacts with no reconstruction: $T_\text{ingest}[\text{wiki}] = 904{,}884{,}660$ against $T_\text{ingest}[\text{rag}] = 1{,}401{,}996$. \textbf{\Hthreea is supported}, by a factor of 645. Nothing about prompt caching changes how many tokens the artifact records; caching changes what those tokens cost.

\paragraph{Post-hoc economic sensitivity.} Because the wiki figure is cache-heavy, a raw token count overstates its economic weight relative to RAG's uncached ingest, and a reader may reasonably ask whether the registered verdict is an artefact of that. It is not. The wiki artifact's \texttt{tokens\_in} aggregates uncached input, cache writes, and cache reads, which bill at $1\times$, $1.25\times$, and roughly $0.1\times$ of base input rate; the per-session split recovered from the build record shows cache reads dominate at $864.2$M of $898.6$M, or $96.2\%$. Re-weighting the input side gives billable-input-equivalents of $129.4$M at the 5-minute cache-write rate and $155.2$M at the 1-hour rate (the time-to-live in effect was not recorded, so we report the range rather than assume a value), and $T_\text{ingest}[\text{wiki}]$ of $135.7$M to $161.5$M once the $6.29$M output tokens are added. \textbf{Output tokens are carried at face value, not weighted}, so these are not pure billable-cost figures either; output is $4.6\%$ of the corrected total and weighting it changes no conclusion. \Hthreea holds on every reading, at a factor of 97 to 115 after weighting against 645 before it. Magnitudes are descriptive; the registered rule is directional. The \Hthree verdict does not depend on any of this: the crossover denominator is the per-query differential, which is negative, so no positive ingest cost produces a positive crossover. \Hthree remains refuted on \Hthreeb, which is conjunctively sufficient.

\subsection{Robustness checks (preregistered)}

\begin{table}[h]
\caption{Bootstrap percentile CI (10{,}000 resamples) and preregistered Bayesian posterior ($\mu \sim \mathcal{N}(0,4)$, $\sigma \sim \text{HalfNormal}(4)$, NUTS) for each \Hone/\Htwo cell. ``Strong'' marks $P(\mu \ge \theta) \ge 0.95$ per the preregistered Bayesian threshold (\Hone $\theta = +2$; \Htwo $\theta = 0$).}
\label{tab:robustness}
\centering
\small
\begin{tabular}{llrlrll}
\toprule
Test & Reading & mean & bootstrap 95\% CI & posterior $\mu$ & $P(\mu \ge \theta)$ & strong? \\
\midrule
\Hone struct & judge-avg & $+1.625$ & $[+0.75, +2.88]$ & $+1.500$ & $0.283$ & no \\
\Hone mapping & judge-avg & $+6.625$ & $[+4.13, +8.50]$ & $+5.589$ & $\mathbf{0.967}$ & \textbf{yes} \\
\Hone struct & primary & $+2.000$ & $[+1.00, +3.25]$ & $+1.819$ & $0.440$ & no \\
\Hone mapping & primary & $+6.250$ & $[+3.50, +8.50]$ & $+5.060$ & $0.938$ & close \\
\Htwo ground & judge-avg & $+0.667$ & $[-1.50, +2.00]$ & $+0.522$ & $0.660$ & no \\
\Htwo ground & primary & $+2.000$ & $[+1.00, +3.00]$ & $+1.808$ & $0.937$ & close \\
\bottomrule
\end{tabular}
\end{table}

Only \Hone \texttt{inter\_paper\_mapping} (judge-avg) clears the preregistered ``strongly corroborated'' bar of $P(\mu \ge \theta) \ge 0.95$. The Bayesian posterior shrinks point estimates toward zero (e.g., \Hone primary \texttt{structural\_integrity} sample mean $+2.000$ shrinks to posterior $+1.819$), reflecting the regularising prior at small $n$. The two robustness checks agree on direction but disagree on \emph{how often} the threshold is robustly cleared.

\subsection{Inter-rater reliability: a ceiling-judge finding}
\label{sec:results-irr}

The preregistered IRR rule fired on all four criteria; max-deltas across the expanded $n=13$ secondary coverage are 6 (\texttt{groundedness}), 4 (\texttt{structural\_integrity}), 8 (\texttt{conflict\_awareness}), 6 (\texttt{inter\_paper\_mapping}). Computed instead on the preregistered $n=3$ IRR subset, max-deltas are 6 / 4 / 4 / 3, still triggering on all four criteria.

Max-delta is what the registered adjustment rule keys on, but it is a worst-case statistic that one cell can move and it says nothing about whether the judges rank systems alike. \Cref{tab:irr-agreement} therefore reports conventional agreement summaries over all 26 scored cells per criterion (13 questions $\times$ 2 systems), computed from the deposited judgments with no additional model calls.

\begin{table}[h]
\caption{Inter-judge agreement per criterion, $n=26$ cells (13 questions $\times$ 2 systems). $\rho$ is Spearman, $\tau_b$ Kendall, QWK quadratic-weighted $\kappa$ on the 1--10 scale, ICC(2,1) two-way random effects with absolute agreement. ``Level'' is mean(\gem) $-$ mean(\gpt). ``$\rho_\Delta$'' correlates the two judges' \emph{within-question} wiki$-$RAG differences ($n=13$), i.e.\ agreement after any constant level shift is removed.}
\label{tab:irr-agreement}
\centering
\small
\begin{tabular}{lrrrrrrr}
\toprule
Criterion & $\rho$ & $\tau_b$ & QWK & ICC & MAD & Level & $\rho_\Delta$ \\
\midrule
\texttt{groundedness}         & 0.035 & 0.020 & 0.039 & 0.041 & 2.96 & $+2.27$ & 0.323 \\
\texttt{conflict\_awareness}   & 0.406 & 0.275 & 0.239 & 0.247 & 2.04 & $+1.50$ & 0.302 \\
\texttt{structural\_integrity} & 0.653 & 0.471 & 0.443 & 0.452 & 1.35 & $+1.12$ & 0.312 \\
\texttt{inter\_paper\_mapping}  & \textbf{0.809} & \textbf{0.610} & \textbf{0.806} & \textbf{0.812} & \textbf{1.27} & $\mathbf{+0.19}$ & \textbf{0.705} \\
\bottomrule
\end{tabular}
\end{table}

Inter-judge agreement varied sharply by criterion, and the ordering is the same on every statistic: highest on the most operationally concrete criterion and lowest on the most holistic one. \texttt{inter\_paper\_mapping}, the one criterion with a countable definition (synthesis across $\ge 2$ papers), reaches $\rho = 0.809$ and ICC $= 0.812$ with a level shift of only $+0.19$: the two judges substantially agree about both the level and the ordering. \texttt{groundedness}, the most holistic criterion, reaches $\rho = 0.035$ and ICC $= 0.041$ against a level shift of $+2.27$.

\textbf{This corrects a weaker claim we made previously.} An earlier version of this analysis, reasoning from max-deltas and criterion-level means, described the judges' disagreement as ``largely about level, not ranking.'' That is true of \texttt{inter\_paper\_mapping} and defensible for \texttt{structural\_integrity} ($\rho = 0.653$), but it is not true of \texttt{groundedness}, where the rank correlation is indistinguishable from zero and removing the level shift does not rescue it ($\rho_\Delta = 0.323$). On the criterion that carries \Htwo, the two judges are not scoring a shared underlying quantity at different heights: they show essentially no rank agreement on this sample. The direction-flip already reported at the criterion level (\gpt rates RAG higher, 7.00 vs 6.31; \gem rates wiki higher, 8.15 vs 9.69) is the visible symptom of that, and the \Htwo subset shows it sharply: primary $\Delta = +2.000$, secondary $\Delta = -0.667$, judge-avg $\Delta = +0.667$, supported under the registered rule though the secondary judge alone would refute.

The per-judge per-system means (\cref{app:irr}) show the mechanism: \gem ratings on the wiki output saturate near 10/10 on all four criteria, while \gpt spreads between 6 and 9 with greater variance. \texttt{inter\_paper\_mapping} is again the exception, with the two judges' RAG means differing by 0.15 (5.38 vs 5.23) and their wiki means by 0.54 (9.46 vs 10.00). We surface as a transferable methodological observation: \emph{agreement was highest on the criterion with a concrete operational definition, in both level and rank, and lowest on the most holistic criterion, where it approached zero}, consistent with prior LLM-as-judge literature~\citep{zheng2023mtbench, liu2023geval, wang2024fairevaluators}. This suggests that criterion design, not judge identity alone, can materially affect LLM-judge reliability. We cannot isolate concreteness as the cause: there are only four criteria, concreteness was not manipulated, and score distribution, ceiling behaviour, task difficulty, and how each system expresses the property all co-vary with it. Establishing the causal claim would need the same criterion rewritten at several levels of operational specificity. The practical consequence does not depend on the mechanism: \emph{criterion-level agreement should be measured rather than assuming a judge is uniformly reliable across a rubric}, and it is cheap to measure.

\subsection{Post-hoc claim-level grounding}
\label{sec:results-claim}

We ran a post-hoc claim-level grounding analysis on all 466 atomic claims. \emph{This analysis is not preregistered} and is reported as exploratory robustness on the \Htwo mechanism.

\paragraph{A correction to this analysis.} The version of this paper first reported a large claim-level advantage for the wiki: its cited claims were scored supported 40.2\% of the time against 18.9\% for RAG, and unsupported 6.2\% against 34.1\%. That result came from an error in the scoring script, which cut every cited passage to its first 1{,}500 characters before the scorer saw it (\cref{sec:experiments}). The cut reached 128 of the 132 cited RAG claims and none of the wiki's, so RAG claims were judged against partial evidence and wiki claims against complete evidence. Rescored on full passages with the same prompt, scorer model, and claims, 57 of the 132 RAG verdicts change. A capped rerun of the same RAG requests changes 19 and leaves the rates near the originals, at 22.7\% supported and 36.4\% unsupported, so the shift comes from the cut and not from rerun noise. All figures below are the corrected ones. \Cref{app:correction} gives the originals beside them, and the statements that rested on the originals are withdrawn.

\paragraph{Important interpretive caveat.} For RAG claims, the cited evidence is the original PDF chunk that the retrieval system surfaced; for wiki claims, the cited evidence is a wiki-page excerpt, which is itself a compilation of source PDFs and not an original PDF passage. The analysis below therefore tests \emph{evidence-artifact claim alignment} (does each claim follow from what the system cited?), \emph{not original-source fidelity} (does each claim follow from the underlying PDF?). A reader who wants source fidelity for wiki must additionally ask: does the wiki page itself accurately reflect the source PDF? We do not measure this; \cref{sec:limitations} flags it as the primary follow-up.

\paragraph{Aggregate by system.} Total claims: RAG 150, Wiki 316. Cited rate (claims with non-empty \texttt{cited\_source\_idx}): RAG 88.0\%, Wiki 76.3\%. Wiki produces about twice as many claims per answer (24.3 vs 11.5) and cites a smaller fraction of them.

\paragraph{Answer length is an uncontrolled confound.} Length here means the answer the judges read, measured on the answer text rather than on token telemetry. Per answer, single-round RAG writes 385 words (2{,}638 characters), decomp-RAG 604 (4{,}185), and the wiki 740 (5{,}250), so the wiki's answers are $1.92\times$ single-round RAG's by word count and $1.99\times$ by character count. Claims per answer run 11.5 / 17.8 / 24.3 and cited claims per answer 10.2 / 17.2 / 18.5 across the same three arms. Pipeline generation tokens (1{,}140 / 5{,}143 / 7{,}472) are a much larger spread and are \emph{not} a length measure: they count every call a system makes, including the wiki's browsing turns, decomp-RAG's per-sub-question generations, and thinking tokens, none of which a judge sees.

Length is not controlled anywhere in this study and plausibly inflates the two rubric criteria the wiki wins on, since a longer answer has more room to supply structure and cross-paper links.

What we will not claim is that the ablation rules length out. Answer length increases monotonically across the three arms in the same order as synthesis performance, and decomp-RAG reaches 82\% of the wiki's word count while closing roughly 94\% of the synthesis gap, which is as consistent with a length effect as against one. We treat length, and the prose-vs-bullet style difference it travels with, as an uncontrolled confound; a length- or retrieval-budget-matched replication is the cheapest experiment that would separate them.

Treating uncited claims as \texttt{unsupported} (\emph{all-claims} view), the rates are: RAG 47.3\% supported / 22.7\% unsupported; Wiki 31.6\% / 29.1\%. The conditional-on-cited view (\emph{citation-precision} sense) is:

\begin{table}[h]
\caption{Claim-level verdicts, conditional on the claim carrying a citation, scored on full passages. Claims were atomized by \opus and each scored by \gpt against the evidence its own answer cited; this axis is single-scorer, unlike the rubric scoring, which carries two-judge IRR coverage. \Cref{app:correction} gives the figures as first reported, from truncated evidence.}
\label{tab:claim-verdicts}
\centering
\small
\begin{tabular}{lrr}
\toprule
Verdict (conditional on cited) & RAG ($n=132$) & Wiki ($n=241$) \\
\midrule
supported & $\mathbf{53.8\%}$ (macro $51.1\%$) & $\mathbf{41.5\%}$ (macro $37.9\%$) \\
partial & 31.1\% & 51.0\% \\
contradicted & 3.0\% & 0.4\% \\
\textbf{unsupported} & $\mathbf{12.1\%}$ (macro $12.1\%$) & $\mathbf{7.1\%}$ (macro $8.7\%$) \\
\bottomrule
\end{tabular}
\end{table}

Macro-averages by question, which weight each of the 13 answers equally regardless of claim count, agree with the micro-averages: RAG's cited claims are scored supported more often than the wiki's, and unsupported somewhat more often.

\paragraph{Human check.} Two annotators who are not authors scored a sample of 100 cited claims, 50 per system, stratified by tier, under a rule fixed in advance: the scorer counts as validated only if its binary supported-versus-not agreement with the annotators reaches 70\% and Cohen's $\kappa$ reaches 0.40 (\cref{app:correction}). The annotators agree with each other on 94 of the 100 claims ($\kappa = 0.81$). The scorer as first run agrees with them on 54 and 50 ($\kappa = 0.23$ and $0.19$), and the corrected scorer on 69 and 65 ($\kappa = 0.40$ and $0.33$). It does not meet the rule against either annotator, before or after the correction. Agreement with an adjudicated consensus, which the design called for, was not assessed, and for the corrected scorer it could reach the 70\% floor (\cref{app:correction}). Every remaining disagreement is a claim the annotators call supported and the scorer calls partial or, in a few cases, contradicted. On the annotators' labels, cited claims are supported 78\% of the time for RAG and 78\% or 86\% for the wiki, and unsupported 16\% of the time for RAG against 6\% or 10\% for the wiki. The difference in unsupported rates is not significant on either annotator's labels by a conventional pooled two-proportion test ($p = 0.11$ and $0.37$). The test fixed in our analysis script before any label existed used unpooled variances with a finite-population correction for the 132 and 241 cited claims sampled, giving $p = 0.049$ and $0.28$, but that correction supports inference about these claims rather than about the systems. Both tests treat claims as independent, although claims cluster within answers, so both overstate precision. The annotators' disagreements were not adjudicated, so each is reported separately.

Of the 5 claims scored \texttt{contradicted}, 4 are RAG (T2-fair-use: a position the cited passage explicitly disclaims; T4-cross-domain: the direction of feature drift, on which the two cited passages themselves disagree; T5-glp1: a false description of a paper's content; B3: a defense's effect inverted) and 1 is Wiki (T4-eelgrass: a $p$-value given as 0.057 where the passage has 0.066). The first version's example of a RAG error on B1, a hazard ratio of 1.38 said to contradict a cited 1.37, was an error by the scorer: the cited table reports 1.37 for the expanded four-point MACE definition and 1.38 for three-point MACE, and the answer reported both correctly.

\paragraph{By tier.} \Cref{tab:claim-tiers} in \cref{app:correction} gives rates by tier for all three arms. RAG's cited claims are scored supported more often than the wiki's on five of the six tiers; the wiki leads only on bias-check, 51.9\% against 36.8\%, on 27 and 19 claims. The two grounding measures agree on multi-hop, where the rubric put RAG 3.50 points ahead on \texttt{groundedness} and RAG's supported rate is 76.0\% against 40.0\%, and they diverge on bias-check, where the rubric put RAG 2.00 points ahead. With one tier agreeing and one diverging, on small numbers of claims, we report the tier pattern as descriptive only. Disagreement between the two measures need not indicate an architectural mechanism, because they measure different things. Rubric \texttt{groundedness} is a holistic rating of a whole answer. The claim-level analysis scores extracted claims against the evidence each cites, and its cited-only rates exclude uncited claims and weight each answer by its number of claims. The wiki's cited evidence is itself an intermediate representation rather than the source paper. And the two measures differ in judge calibration, claim extraction, and how evidence is presented to the scorer.

\emph{Absolute-rate caveat.} These rates should not be read as high absolute citation reliability for either system. The wiki's cited claims are more often \texttt{partial} than \texttt{supported} (51.0\% against 41.5\%), and the human check shows the scorer draws the supported-partial boundary more strictly than people do: the annotators called 78 and 82 of the 100 sampled claims supported, where the scorer as first run called 32. The result is a relative comparison under a strict boundary, not a benchmark-level claim of citation precision.

\paragraph{Reading.} What the corrected analysis supports is narrow. The corrected scorer rates RAG's cited claims supported more often than the wiki's, and the human sample does not establish a difference in either direction. Both find RAG's cited claims unsupported somewhat more often, by about 1.7 times on the corrected scorer (12.1\% against 7.1\%) and 1.6 to 2.7 times on the annotators' labels, a difference that is not significant on either annotator's labels by a conventional test. The version first reported found a wiki advantage of about two times on supported claims and four to five times on unsupported ones, and proposed that wiki compilation pre-positions evidence in claim-shaped artifacts. That result came from the truncation, and the mechanism proposed for it is withdrawn.

\textbf{The preregistered \Htwo verdict is unaffected.} It rests on rubric \texttt{groundedness}, whose judges received the full cited passages. The first version argued that the claim-level analysis reversed the mechanism behind \Htwo, with RAG showing worse evidence-artifact alignment than the wiki; that argument rested on the truncated scoring and is withdrawn.

The methodological lesson we draw concerns the instrument. The claim-level scorer was consistent with itself, changing 10\% to 14\% of verdicts on a rerun, yet it agreed with people on only half the claims, and the largest cause was an input defect that no consistency check could have detected. Automated grounding scores should be checked against human labels on a sample before they are used to compare systems, and agreement between two models should not be read as agreement with people.

\subsection{Other preregistered exploratory analyses}

The preregistration listed seven secondary analyses planned but not adjudicative. We summarize here; full data in \cref{app:exploratory}.
\begin{itemize}
\item \emph{Bias-tag stratification.} The question taxonomy holds: bias=rag questions show RAG winning \texttt{groundedness} by 2.0; bias=wiki questions show Wiki ahead on three criteria (largest on \texttt{inter\_paper\_mapping} at $+5.75$) and slightly behind on \texttt{groundedness} ($-0.25$).
\item \emph{Latency.} RAG total 3.3 min, Wiki total 22.0 min ($6.6\times$); the wall-clock ratio is smaller than the token ratio.
\item \emph{Adaptive-thinking engagement.} RAG triggered nonzero \texttt{thinking\_tokens} on 2/13 questions; Wiki on 13/13. Both at the same \texttt{xhigh} flag, so the flag does not account for the difference; what does was not measured.
\item \emph{Cost-vs-quality scatter.} RAG cluster ($\le 8$k tokens, scores 3.75--8.25) and Wiki cluster ($\sim 30$k--210k tokens, 6.75--9.25) are not globally dominated. On 4 of 13 questions RAG matches or beats Wiki on rubric mean at an order-of-magnitude lower cost (T1-cardio, T3-rwd, B2, B3). The Pareto frontier is task-dependent.
\end{itemize}

\section{Exploratory decomp-RAG ablation}
\label{sec:ablation}

A natural alternative hypothesis is that single-round retrieval is a known weakness on multi-hop synthesis~\citep{trivedi2023ircot, jiang2023flare}, so wiki's H1 advantage might be partly an artefact of the registered RAG configuration. We test this with a post-hoc decomposition-retrieval ablation.

\paragraph{Method.} For each question, \opus decomposes it once, not iteratively, into 2--5 self-contained sub-questions returned as JSON (the prompt is verbatim in \cref{app:decomp-tiers}); point-source questions still receive two or three sub-questions targeting adjacent context, and no separate step validated the sub-questions; the existing single-round RAG pipeline (\cref{sec:methods}) runs per sub-question, keeping cited sources only; cited chunks are deduplicated; \opus generates the final answer using the \emph{same} answer-generation system prompt as the single-round baseline (verbatim from the codebase). Same query model, corpus, vector store, and reranker; only the question-to-retrieval-context step changes. Three predictions were committed before running: \emph{P1} decomp closes $\ge 50\%$ of the wiki--single-RAG gap on \texttt{inter\_paper\_mapping}; \emph{P2} per-query token cost between $5$--$10\times$ single-round; \emph{P3} claim-level supported-rate improves on multi-hop. Both judges were re-invoked on the (decomp, wiki) pair with the same prompts, rubric, and seeds as the main run: \gpt primary at \texttt{seed=42}, \gem secondary at \texttt{seed=43}, \texttt{medium} reasoning effort, full 13-question coverage. \textbf{The preregistered max-delta IRR trigger fires on all four criteria here too} (max-deltas 6 / 4 / 4 / 4 on \texttt{groundedness} / \texttt{structural\_integrity} / \texttt{conflict\_awareness} / \texttt{inter\_paper\_mapping}), so we report the ablation under the same \textbf{primary-only} and \textbf{IRR-adjusted} readings used in \cref{sec:results}, with the judge-average as the headline. Applying the registered adjustment rule keeps the ablation on the same measurement scale as the confirmatory analysis; it does not make it a preregistered re-test, and \emph{the results below remain exploratory and judge-sensitive}.

\paragraph{Three-way result.}

\begin{table}[h]
\caption{Three-way comparison summary. Cost rows sum across the 13 questions. Synthesis-shape rows use the \Hone confirmatory subset ($n=4$) and are reported under both readings: \emph{primary} is \gpt alone, \emph{judge-avg} is the IRR-adjusted mean of the two judges. Each gap is computed \emph{within} a single judge run, so the Wiki column gives the wiki mean from the run the adjacent RAG arm was scored in; the two runs score identical wiki answers and agree exactly except on \texttt{structural\_integrity} under judge-avg, where both values are shown as original\,/\,decomp. Claim-level row uses cited claims across all 13 questions, scored on full passages (corrected; \cref{app:correction}), and is single-scorer on every arm. $^a$Single-RAG, scored in the original judge run. $^b$Decomp-RAG, scored in the post-hoc decomp judge run with the same prompts, rubric, and seeds.}
\label{tab:ablation-summary}
\centering
\small
\begin{tabular}{@{}llrrrp{3.0cm}@{}}
\toprule
Metric & Reading & Single$^a$ & \textbf{Decomp$^b$} & Wiki & Wiki gap, single $\to$ decomp \\
\midrule
Total query tokens & --- & 78k & \textbf{491k} & 1{,}651k & decomp $3.4\times$ cheaper than wiki on LLM tokens \\
Total wall-clock (min) & --- & 3.3 & \textbf{26.8} & 22.0 & decomp slower than wiki (sequential) \\
\midrule
\Hone \texttt{inter\_paper\_mapping} & primary & 3.50 & \textbf{9.00} & 9.75 & gap $+6.25 \to +0.75$ ($88\%$ closed) \\
 & judge-avg & 3.25 & \textbf{9.50} & 9.88 & gap $+6.62 \to +0.38$ ($94\%$ closed) \\
\Hone \texttt{structural\_integrity} & primary & 7.00 & \textbf{8.75} & 9.00 & gap $+2.00 \to +0.25$ ($88\%$ closed) \\
 & judge-avg & 7.88 & \textbf{9.38} & 9.50\,/\,9.12 & gap $+1.62 \to -0.25$ (decomp edges above wiki) \\
\midrule
Cited claim supported \% & single-scorer & 53.8 & 32.4 & 41.5 & decomp lowest of the three arms \\
\bottomrule
\end{tabular}
\end{table}

\textbf{Cost (P2).} Decomp's per-query LLM-token cost (under the H3 token-accounting scope, excluding reranker and embedding spend) is $6.3\times$ single-round and $3.4\times$ cheaper than wiki, satisfying P2. Wall-clock is \emph{slower} than wiki despite fewer tokens, because decomp performs sequential retrieval-then-rerank-then-validate calls per sub-question; the LLM-token picture is favorable for decomp, the wall-clock picture is not, and the embedding/reranker cost picture is not measured in this study.

\textbf{Synthesis structure (P1).} On the \Hone confirmatory subset, the decomp run supports P1 under both readings, and more strongly under the headline one. On \texttt{inter\_paper\_mapping} the wiki advantage shrinks from $+6.62$ to $+0.38$ judge-averaged, about $94\%$ of the gap closed, and from $+6.25$ to $+0.75$ primary-only, about $88\%$. On \texttt{structural\_integrity} it shrinks from $+1.62$ to $-0.25$ judge-averaged, the sign indicating that decomp edges slightly \emph{above} the wiki, and from $+2.00$ to $+0.25$ primary-only. Against the decomp baseline, both criteria fall below the preregistered $+2.0$ threshold under either reading; the wiki would not clear what the registered \Hone thresholds required in this comparison. \emph{This does not alter the preregistered \Hone verdict against single-round RAG. It indicates that the synthesis-structure advantage is sensitive to the RAG baseline.}

\textbf{Holistic groundedness, and a second instance of the ceiling-judge finding.} The two judges disagree by direction on holistic \texttt{groundedness} in the ablation, exactly as they do in the main run (\cref{sec:results-irr}). \gpt scores decomp above the wiki by $+1.15$ across all 13 questions, \gem scores the wiki above decomp by $-0.85$, and the judge-average is $+0.15$, a small observed difference that we do not read as a decomp advantage; no equivalence analysis was run, so it does not establish parity either. The same pattern holds on the tiers: $+1.75$ judge-averaged on multi-hop against $+3.00$ primary-only, and $+0.67$ on the \Htwo bias-check parity test against $+2.33$ primary-only, still meeting the registered \Htwo criterion but with a much smaller margin. \texttt{groundedness} carries the largest inter-judge disagreement in the ablation as well (mean $|\Delta| = 2.54$ against $0.92$ for \texttt{inter\_paper\_mapping}), reproducing the main run's finding on an independent set of answers: \emph{holistic groundedness is the criterion least robust to judge calibration, and a single-judge reading of it can invert.}

\textbf{Claim-level grounding (P3 not supported).} We re-ran the §5.4 atomize-and-score pipeline, which is single-scorer (\gpt) on every arm, on the 232 decomp claims (94.4\% cited, vs 88.0\% single, 76.3\% wiki). Scored on full passages, decomp's cited-claim supported rate is 32.4\%, below both single-round RAG (53.8\%) and the wiki (41.5\%), and its unsupported rate, 32.0\%, is the highest of the three. Decomp is ahead of single-round RAG on one tier, bias-check (60.9\% against 36.8\%). On multi-hop, the tier P3 named, it is well behind (39.1\% against 76.0\%), so P3 is not supported. The version first reported put decomp's supported rate at 19.2\% and read P3 as partially supported; both figures came from the same truncation, which cut 200 of decomp's 219 cited claims (\cref{app:correction}). The human check did not include decomp claims.

\paragraph{Reading.} When RAG breaks each question into sub-questions, it closes about $94\%$ of the \texttt{inter\_paper\_mapping} gap under the adjusted reading, so the decomposition variant substantially reduced the observed synthesis-score gap. Answer length rises across the three arms in the same order as synthesis scores and is not controlled (\cref{sec:results-claim}), so the result is also consistent with a length effect. The first version of this paper argued that a separate claim-level advantage remained for the wiki, attributable to representation alignment; that argument rested on the truncated scoring and is withdrawn, so the corrected citation analysis no longer supplies evidence for that mechanism. The study does not isolate compilation, traversal, retrieval coverage, or answer length well enough to establish their separate causal contributions. On the corrected scorer, decomposition's broader retrieval came with lower citation support, not higher.

\paragraph{Caveats summary.} Two-judge with the registered IRR adjustment applied, but the rubric scoring inherits the calibration drift documented in \cref{sec:results-irr}, which is direction-flipping on holistic \texttt{groundedness} here as well; the claim-level pipeline remains single-scorer on all three arms, and its human check (\cref{app:correction}) covers single-round RAG and wiki claims only. Cross-run wiki absolute scores drift by at most $0.62$ on any criterion's 13-question mean under either judge and at most $0.15$ under the judge-average; on the two criteria used in the \Hone rows the drift is $0.00$ on \texttt{inter\_paper\_mapping} under both judges and at most $0.75$ on \texttt{structural\_integrity} under \gem, which is why \cref{tab:ablation-summary} computes every gap within a single judge run. $n=4$ on the \Hone subset. Because the ablation is exploratory and small-$n$, we report point estimates only and treat them as mechanism evidence rather than confirmatory inference; one decomposition prompt was fixed before the run. Reproducer: \texttt{experiments/run\_decomp\_rag.py} + \texttt{run\_claim\_grounding\_decomp.py} + \texttt{analyze\_decomp\_dual\_judge.py}.

\section{Discussion and limitations}
\label{sec:discussion}
\label{sec:limitations}

\paragraph{What the findings mean.} The wiki's advantage in this comparison is an advantage in synthesis structure against a single-round baseline, measured by LLM judges, and decomposition retrieval reduced most of the observed gap (\cref{sec:ablation}). Because the main comparison changes corpus representation, retrieval strategy, and test-time compute together, the part that remains cannot be assigned to the compiled representation. The first version of this paper proposed a second, \emph{representation-alignment} mechanism, resting on a claim-level citation advantage for the wiki. That advantage came from a scoring error and does not survive correction or the human check (\cref{sec:results-claim}), and the proposal is withdrawn.

Karpathy's three intuitions are partially borne out. (i) Pre-compilation may help with synthesis (the compiled-wiki configuration scored higher than the registered single-round baseline on LLM-judged criteria, and the decomposition variant substantially reduced that gap). (ii) Pre-compilation may sacrifice point-source fidelity (supported on the registered rubric test; the corrected claim-level analysis is mixed, with RAG's cited claims scored supported more often and also unsupported somewhat more often, and we do not measure original-PDF fidelity). (iii) Pre-compilation shifts cost from query to ingest with finite-N amortization (refuted: per-query goes \emph{up}, not down).

\paragraph{Tradeoffs observed in these configurations.} These are observations about the three configurations tested, on one model and one corpus, not deployment advice. Single-round RAG used the fewest tokens and met the registered \Htwo test on point-source groundedness, a judge-sensitive result. Decomp-RAG scored close to the wiki on multi-paper synthesis at about $6\times$ single-round LLM-token use, excluding the additional reranker and embedding spend from its sub-question retrievals; it was tested at one operating point, and on the corrected claim-level scorer its citations support its claims less often than either alternative's. We withdraw the first version's recommendation of the wiki for claim-citation alignment: on corrected scoring the wiki's cited claims are supported less often than single-round RAG's, and the annotators' labels did not establish better support for either. In this configuration the case for the wiki rests on synthesis structure against a single-round baseline, at a $\sim 21\times$ per-query token premium. The defensible form of \Hthree's per-query result is narrow: \emph{the tool-browsing wiki configuration tested here cannot amortize its ingest cost through query-token savings, because its measured per-query token use exceeds the tested RAG baseline's rather than falling below it.} It does not generalize to compiled-knowledge architectures in general. \citet{zhang2026policywiki} report the opposite direction on query-time context with a wiki served as retrieved pages, and \citet{chan2024cag} and \citet{huerta2026wicer} serve compiled knowledge from a cached context window, which prices traversal out of the query path entirely. What we measured is the per-query cost of a configuration in which a tool-using agent browses the wiki; the study does not separate the contributions of traversal and compilation to that cost.

\paragraph{What the findings do not show.} \emph{Limited human evaluation} (the most consequential gap). A human check covers the claim-level scorer on a sample of 100 single-round RAG and wiki claims (\cref{app:correction}), and the scorer did not meet the agreement rule fixed in advance against either annotator, before or after correction; agreement with an adjudicated consensus was not assessed. The rubric judges and the decomposition arm's claim-level scores have no human check, so the \Hone and \Htwo results are comparisons of LLM-judged scores (\Hthree rests on token telemetry), and the IRR finding (\cref{sec:results-irr}) shows that LLM-judge calibration drift is large and direction-flipping on holistic criteria. Agreement between the two judges measures consistency between models, not agreement with people. The §6 decomp ablation now carries two-judge coverage and the same registered IRR adjustment as the confirmatory analysis, but it should still be read as exploratory and judge-sensitive: its holistic \texttt{groundedness} comparison reverses direction between the two judges, and its claim-level component remains single-scorer. \emph{Small $n$} ($n=4$ for \Hone, $n=3$ for \Htwo): standard errors are large and the decomp ablation reports point estimates only. Preregistration improves the credibility of the analysis but does not address statistical power or representativeness, and every conclusion is limited to the tested corpus, questions, and system configurations. \emph{Single corpus and single model}: 24 papers, and \opus performs every generating step, answering for both systems, compiling the wiki, enriching the RAG index, and atomizing claims; transfer to larger corpora or other model families is unverified, and the wiki's results depend on its compile model as well as its query model. \emph{Style and length confound}: wiki produces continuous prose, RAG bullet-and-cite, and the wiki's answers are $1.9\times$ longer in visible text (740 vs 385 words per answer) with 24.3 claims against 11.5; neither is controlled for, and both plausibly favor the criteria the wiki wins on (\cref{sec:results-claim}). \emph{One decomposer prompt} fixed before the run, not iterated. \emph{Cost telemetry under prompt caching}: the registered \Hthreea verdict rests on the deposited artifacts alone, but the deposited aggregate does not license a \emph{billable-cost} reading (\cref{sec:experiments}), a caveat its manifest records at deposit time. The per-session components reconcile exactly to that aggregate, so the quantity is recoverable and we report it in \cref{sec:results}; the component table and a reconstruction script ship in the supplementary material, and the script refuses to report a figure unless the components sum to the deposited \texttt{totals.tokens\_in} to the token. Three residual limits: the cache time-to-live was not recorded, so the write rate is bounded rather than known and we report both endpoints; output tokens are carried unweighted, so the corrected $T_\text{ingest}$ is not a pure billable-cost figure; and the artifact carries no per-paper attribution of ingest tokens. None affects the \Hthreea direction, which holds by roughly two orders of magnitude, or \Hthreeb, where query-side caching was disabled on both sides. \emph{Methods note for replicators:} cached-vs-uncached cost-comparison studies must capture four columns per LLM call (uncached / cache\_creation / cache\_read / output), not a rolled-up \texttt{tokens\_in}.

\paragraph{Beyond this study.} Future work should compare claim-level grounding against original PDF passages (testing source fidelity for wiki) rather than against the cited evidence artifact only; extend claim-level scoring to a second scorer, the one axis of this study still resting on a single model; adjudicate the human annotation and extend it to the rubric criteria and the decomposition arm; test retrieval-during-generation methods~\citep{trivedi2023ircot, jiang2023flare} that may close more of the wiki advantage; separate the answer-length confound with a length- or retrieval-budget-matched replication, which would also test whether decomp-RAG's synthesis gains are a chunk-exposure effect rather than a decomposition effect; and test hybrid wiki+source-RAG architectures that could backstop wiki's rare paraphrase artefacts (1 of 5 contradicted claims) with source-chunk verification.

\section{Conclusion}
\label{sec:conclusion}

The most defensible conclusion of this preregistered comparison is not that compiled wiki memory beats RAG, nor the reverse. On one model and one small corpus, the compiled wiki organized multi-paper evidence better than single-round retrieval on LLM-judged rubric criteria, at about $21\times$ the per-query LLM-token use. A decomposition-retrieval baseline came within $0.4$ points of it on both criteria under the judge-averaged reading, at $\sim 3.4\times$ lower per-query LLM-token use. On claim-level citation support, once a scoring error was corrected and a sample was checked against human annotators, the analysis did not establish an advantage for the wiki over single-round RAG. Evaluations of this kind should report synthesis structure, citation support, and cost separately, and should check any automated grounding scorer against human labels before comparing systems with it.

\bibliography{bibliography}

\appendix
\section{Per-question rubric scores (primary judge)}
\label{app:per-question}

Primary-judge ratings for all 13 questions, two systems, and four criteria, 104 scores in all. Order: \texttt{groundedness} / \texttt{structural\_integrity} / \texttt{conflict\_awareness} / \texttt{inter\_paper\_mapping}, all on a 1--10 anchored scale.

\begin{table}[h]
\centering
\small
\begin{tabular}{llrrrr|rrrr}
\toprule
\multicolumn{2}{c}{} & \multicolumn{4}{c}{RAG (g/s/c/m)} & \multicolumn{4}{c}{Wiki (g/s/c/m)} \\
\cmidrule(lr){3-6}\cmidrule(lr){7-10}
qid & tier & g & s & c & m & g & s & c & m \\
\midrule
T1-cardio-baselines & chronological & 9 & 7 & 8 & 7 & 4 & 9 & 7 & 10 \\
T1-mia-evolution & chronological & 3 & 6 & 4 & 6 & 7 & 10 & 9 & 10 \\
T2-fair-use-vs-optin & conflict & 7 & 8 & 6 & 7 & 8 & 9 & 10 & 10 \\
T2-oae-bio-vs-chemical & conflict & 4 & 6 & 2 & 3 & 7 & 10 & 9 & 10 \\
T3-mia-as-copyright-evidence & multi-hop & 8 & 7 & 8 & 2 & 6 & 9 & 9 & 10 \\
T3-rwd-validity-for-side-effects & multi-hop & 9 & 8 & 6 & 7 & 4 & 9 & 5 & 9 \\
T4-cross-domain-audit & emergence & 8 & 8 & 6 & 4 & 5 & 9 & 8 & 10 \\
T4-eelgrass-mrv-gaps & emergence & 7 & 5 & 8 & 1 & 8 & 9 & 9 & 10 \\
T5-glp1-policy & policy & 7 & 7 & 7 & 3 & 8 & 9 & 8 & 10 \\
T5-25th-paper & policy & 3 & 8 & 6 & 6 & 5 & 9 & 8 & 10 \\
B1-devote3-confidence-interval & bias-check & 9 & 6 & 9 & 7 & 8 & 9 & 9 & 8 \\
B2-he-tyka-equilibration-ratios & bias-check & 9 & 6 & 9 & 9 & 6 & 8 & 9 & 9 \\
B3-mia-attack-success-rate & bias-check & 8 & 7 & 9 & 8 & 6 & 7 & 9 & 7 \\
\bottomrule
\end{tabular}
\end{table}

Per-tier means (primary judge) with $\Delta$ = wiki $-$ rag:

\begin{table}[h]
\centering
\small
\begin{tabular}{lrrrrr}
\toprule
Tier & $n$ & $\Delta$\textsubscript{ground} & $\Delta$\textsubscript{struct} & $\Delta$\textsubscript{conflict} & $\Delta$\textsubscript{mapping} \\
\midrule
bias-check & 3 & $\boldsymbol{-2.00}$ & $+1.67$ & $+0.00$ & $+0.00$ \\
chronological & 2 & $-0.50$ & $+3.00$ & $+2.00$ & $+3.50$ \\
conflict & 2 & $+2.00$ & $+2.50$ & $+5.50$ & $+5.00$ \\
multi-hop & 2 & $\boldsymbol{-3.50}$ & $+1.50$ & $+0.00$ & $+5.00$ \\
emergence & 2 & $-1.00$ & $+2.50$ & $+1.50$ & $+7.50$ \\
policy & 2 & $+1.50$ & $+1.50$ & $+1.50$ & $+5.50$ \\
\bottomrule
\end{tabular}
\end{table}

\paragraph{\Hone{} confirmatory subset: paired judge diffs.} Wiki $-$ RAG on the multi-hop + emergence subset, per judge and judge-averaged. Moved here from \cref{sec:results} for space; the means are quoted inline there.

\begin{table}[h]
\centering
\small
\begin{tabular}{lrrrrrr}
\toprule
& \multicolumn{3}{c}{\texttt{structural\_integrity}} & \multicolumn{3}{c}{\texttt{inter\_paper\_mapping}} \\
\cmidrule(lr){2-4} \cmidrule(lr){5-7}
Question & primary & secondary & avg & primary & secondary & avg \\
\midrule
T3-mia-as-copyright-evidence & $+2$ & $0$ & $+1.0$ & $+8$ & $+7$ & $+7.5$ \\
T3-rwd-validity-for-side-effects & $+1$ & $0$ & $+0.5$ & $+2$ & $+4$ & $+3.0$ \\
T4-cross-domain-audit & $+1$ & $+2$ & $+1.5$ & $+6$ & $+8$ & $+7.0$ \\
T4-eelgrass-mrv-gaps & $+4$ & $+3$ & $+3.5$ & $+9$ & $+9$ & $+9.0$ \\
\midrule
\textbf{mean} & $\boldsymbol{+2.000}$ & $+1.250$ & $\boldsymbol{+1.625}$ & $\boldsymbol{+6.250}$ & $+7.000$ & $\boldsymbol{+6.625}$ \\
\bottomrule
\end{tabular}
\end{table}

\section{Bayesian model convergence diagnostics}
\label{app:bayes}

The preregistered Bayesian model is fit per criterion $\times$ tier as $\text{score\_diff} \sim \mathcal{N}(\mu, \sigma^2)$ with priors $\mu \sim \mathcal{N}(0, 4)$, $\sigma \sim \text{HalfNormal}(4)$, sampled with NUTS (4 chains $\times$ 2{,}000 iter $\times$ 2{,}000 warmup, target accept 0.95, \texttt{random\_seed=42}). The preregistration's convergence criteria ($\hat R \le 1.01$, ESS $\ge 400$) are met by a wide margin on every cell: $\hat R \le 1.003$ and ESS $\ge 1{,}907$ across all six (test, reading) combinations reported in \cref{tab:robustness}. The Bayesian sidecar reproduces deterministically from the deposited artifacts via \texttt{experiments/run\_pre\-reg\-istered\_analysis.py}.

\section{Inter-rater reliability: full deltas}
\label{app:irr}

\paragraph{Reproducing the agreement statistics.} \Cref{tab:irr-agreement} in the main text is produced by \texttt{analyze\_irr\_stats.py} in the supplementary material, which reads the deposited judgment artifact and makes no model calls. It is dependency-free (no \texttt{numpy}/\texttt{scipy}), so the figures can be checked without building an environment, and it prints a reproduction check against the published claim counts. Ties are handled with average ranks for $\rho$ and $\tau_b$; QWK uses quadratic weights on the full 1--10 scale rather than only observed categories; ICC(2,1) is the two-way random-effects, absolute-agreement, single-measurement form, appropriate here because the two judges are the objects of study rather than interchangeable draws.

\paragraph{Per-question max-deltas.} Computed as $|\text{primary\_score}_{c,q,s} - \text{secondary\_score}_{c,q,s}|$, max over (question, system). Retained because it is the statistic the preregistered adjustment rule keys on, not because it is the best summary of agreement; \cref{tab:irr-agreement} gives the conventional summaries.

\begin{table}[h]
\caption{Per-criterion max-delta under both the preregistered $n=3$ IRR subset and the expanded $n=13$ coverage. Both fire the $> 2$ trigger on all four criteria.}
\label{tab:irr-trigger}
\centering
\small
\begin{tabular}{lrr}
\toprule
Criterion & max-$\Delta$ ($n=13$ expanded) & max-$\Delta$ ($n=3$ prereg subset) \\
\midrule
\texttt{groundedness} & 6 & 6 \\
\texttt{structural\_integrity} & 4 & 4 \\
\texttt{conflict\_awareness} & 8 & 4 \\
\texttt{inter\_paper\_mapping} & 6 & 3 \\
\bottomrule
\end{tabular}
\end{table}

The preregistered $n=3$ IRR subset comprises B1-devote3-confidence-interval (one bias-check), T3-mia-as-copyright-evidence (one multi-hop), and T1-cardio-baselines (one chronological / neutral). Restricting to this subset produces the right column above.

\paragraph{Per-question paired-judge deltas (full $n=13$).} For each question and each system $s \in \{\text{rag}, \text{wiki}\}$, the absolute difference $|p_{c,q,s} - s_{c,q,s}|$:

\begin{table}[h]
\centering
\scriptsize
\begin{tabular}{llrrrrrrrr}
\toprule
\multicolumn{2}{l}{} & \multicolumn{4}{c}{|primary $-$ secondary| RAG} & \multicolumn{4}{c}{|primary $-$ secondary| Wiki} \\
\cmidrule(lr){3-6}\cmidrule(lr){7-10}
qid & tier & g & s & c & m & g & s & c & m \\
\midrule
B1-devote3-confidence-interval & bias-check & 1 & 4 & 1 & 3 & 1 & 1 & 1 & 2 \\
B2-he-tyka-equilibration-ratios & bias-check & 5 & 1 & 1 & 6 & 4 & 2 & 1 & 1 \\
B3-mia-attack-success-rate & bias-check & 2 & 3 & 1 & 2 & 3 & 1 & 1 & 3 \\
T1-cardio-baselines & chronological & 1 & 1 & 2 & 2 & 6 & 1 & 3 & 0 \\
T1-mia-evolution & chronological & 1 & 1 & 1 & 2 & 3 & 0 & 1 & 0 \\
T2-fair-use-vs-optin & conflict & 3 & 1 & 1 & 1 & 2 & 1 & 0 & 0 \\
T2-oae-bio-vs-chemical & conflict & 6 & 3 & 8 & 2 & 3 & 0 & 1 & 0 \\
T3-mia-as-copyright-evidence & multi-hop & 2 & 3 & 2 & 1 & 4 & 1 & 4 & 0 \\
T3-rwd-validity-for-side-effects & multi-hop & 0 & 2 & 2 & 1 & 6 & 1 & 5 & 1 \\
T4-cross-domain-audit & emergence & 2 & 0 & 4 & 2 & 5 & 1 & 2 & 0 \\
T4-eelgrass-mrv-gaps & emergence & 1 & 2 & 2 & 0 & 2 & 1 & 1 & 0 \\
T5-25th-paper & policy & 5 & 0 & 2 & 2 & 5 & 1 & 2 & 0 \\
T5-glp1-policy & policy & 2 & 2 & 2 & 2 & 2 & 1 & 2 & 0 \\
\bottomrule
\end{tabular}
\end{table}

The largest deltas concentrate on \texttt{conflict\_awareness} (max 8 on T2-oae) and \texttt{groundedness} (max 6 on T1-cardio, T3-rwd, T2-oae, B2). The smallest are on \texttt{inter\_paper\_mapping}, where the two judges agree to within 1 point on most cells.

\paragraph{Calibration adjustment effect.} Mean-of-judges per tier means used in \Hone/\Htwo follow the preregistered rule (per-question average across judges, then aggregate by tier). Both readings (primary-only and IRR-adjusted) are reported in \cref{sec:results}.

\paragraph{Per-judge per-system means.} Across all 13 questions. \gem ratings on Wiki sit near 10/10 on all four criteria, highest on \texttt{inter\_paper\_mapping} (10.00). What distinguishes \texttt{inter\_paper\_mapping}, the only criterion with a concrete operational definition, is closer agreement between the judges, the same ordering the agreement statistics in \cref{tab:irr-agreement} recover.

\begin{table}[h]
\centering
\small
\begin{tabular}{lrrrrrrrr}
\toprule
Judge & RAG\textsubscript{g} & wiki\textsubscript{g} & RAG\textsubscript{s} & wiki\textsubscript{s} & RAG\textsubscript{c} & wiki\textsubscript{c} & RAG\textsubscript{m} & wiki\textsubscript{m} \\
\midrule
\gpt (primary) & 7.00 & 6.31 & 6.85 & 8.92 & 6.77 & 8.38 & 5.38 & 9.46 \\
\gem (secondary) & 8.15 & \textbf{9.69} & 8.15 & \textbf{9.85} & 8.54 & \textbf{9.62} & 5.23 & \textbf{10.00} \\
\bottomrule
\end{tabular}
\end{table}

\section{Claim-level grounding: methods, prompts, and limitations}
\label{app:claimgrounding}

\subsection{Two-stage protocol}

\paragraph{Atomization.} For each of the 26 (question $\times$ system) answer cells, we passed the answer plus its numbered source list to \opus (medium adaptive thinking) with the system prompt \emph{``You are a careful factual analyst. You decompose answers into atomic factual claims so they can be checked individually against their cited sources.''} Rules given to the atomizer: each claim must express one specific factual assertion; each must be self-contained (resolve pronouns); paraphrase or quote faithfully; skip pure framing/hedges/meta-commentary; for each claim list the 0-based indices of cited sources, or [] if uncited or meta.

\paragraph{Scoring.} For each claim with non-empty \texttt{cited\_source\_idx}, we passed (claim, concatenated cited \texttt{chunk\_text}) to \gpt (medium reasoning, JSON-mode) with the system prompt \emph{``You are a careful factual evaluator. Given a CLAIM and the cited PASSAGES from the answer's sources, decide whether the claim is supported by the passages.''} Verdicts:
\begin{itemize}
\item \texttt{supported}: every factual element of the claim explicitly stated or unambiguously implied by the passages.
\item \texttt{partial}: some elements supported, others go beyond the passages.
\item \texttt{contradicted}: the passages directly contradict the claim.
\item \texttt{unsupported}: the claim is not addressed by the passages at all, or no passages were cited.
\end{itemize}
The scorer was instructed to be strict on the supported/partial boundary: any element going beyond the passages drops the claim to \texttt{partial}.

Uncited claims (\texttt{cited\_source\_idx = []}) automatically receive \texttt{unsupported}.

\paragraph{Correction.} As first run, the scoring script cut each cited passage to its first 1{,}500 characters before the scorer saw it. The cut reached 128 of the 132 cited RAG claims and none of the 241 cited wiki claims. Every figure in this appendix is from the rescoring on full passages, with the same prompt, scorer model, and claims; \cref{app:correction} gives the first-reported figures beside the corrected ones and a human check of the scorer.

\paragraph{Cost and reproducibility.} Total LLM spend for the full run was approximately \$2 USD. The script supports \texttt{--cached} to reuse cached atomization/scoring outputs; full from-scratch reruns produce different per-claim verdicts (LLM nondeterminism): a rerun of the original requests changed 10\% to 14\% of verdicts and moved aggregate rates by up to four points (\cref{app:correction}).

\subsection{Aggregate rates by tier (cited claims only, corrected)}

\begin{table}[h]
\centering
\small
\begin{tabular}{lrrrrrr}
\toprule
Tier & sys & $n$ & sup\% & part\% & unsup\% & cont\% \\
\midrule
bias-check & rag & 19 & 36.8 & 21.1 & 36.8 & 5.3 \\
bias-check & wiki & 27 & 51.9 & 40.7 & 7.4 & 0.0 \\
chronological & rag & 25 & 56.0 & 24.0 & 20.0 & 0.0 \\
chronological & wiki & 61 & 50.8 & 45.9 & 3.3 & 0.0 \\
conflict & rag & 22 & 63.6 & 31.8 & 0.0 & 4.5 \\
conflict & wiki & 35 & 45.7 & 42.9 & 11.4 & 0.0 \\
emergence & rag & 20 & 50.0 & 25.0 & 20.0 & 5.0 \\
emergence & wiki & 32 & 31.2 & 65.6 & 0.0 & 3.1 \\
multi-hop & rag & 25 & 76.0 & 24.0 & 0.0 & 0.0 \\
multi-hop & wiki & 45 & 40.0 & 48.9 & 11.1 & 0.0 \\
policy & rag & 21 & 33.3 & 61.9 & 0.0 & 4.8 \\
policy & wiki & 41 & 26.8 & 63.4 & 9.8 & 0.0 \\
\bottomrule
\end{tabular}
\end{table}

\subsection{The five contradicted claims}

\begin{table}[h]
\centering
\small
\begin{tabular}{p{1cm}p{4.5cm}p{6cm}}
\toprule
sys & claim (abridged) & scorer reason (abridged) \\
\midrule
RAG & ``Atkinson locates the problem in legal doctrine, arguing that fair use is being over-applied'' (T2-fair-use) & The passage says the paper does not locate the problem in the doctrine itself, but in its use as a broad excuse. \\
RAG & ``memorized samples exhibit smaller, more structured feature drift'' (T4-cross-domain) & One cited passage says smaller drift and the other says stronger; the scorer counts the opposite direction as a contradiction. \\
RAG & ``Source 3 (Martinussen) provides only the title page and authorship'' (T5-glp1) & The passage gives methodological content: a joint QoL-survival framework with debiased estimation. \\
RAG & ``against adversarial regularization, label-only attacks exploit beyond 3 points of the gap attack'' (B3) & The passage says adversarial regularization prevents attacks from exploiting beyond 3 points. \\
Wiki & ``Jones et al.\ (2025) report $R^2=0.316$, $p=0.057$ for isopod mortality'' (T4-eelgrass) & The passage gives $p = 0.066$. \\
\bottomrule
\end{tabular}
\end{table}

The first run scored two other claims contradicted. For B1 it read a hazard ratio of 1.38 against the cited table's 1.37, but 1.37 is the table's figure for the expanded four-point MACE definition, 1.38 is its three-point figure, and the answer reported both correctly. The wiki's T2-oae claim is scored supported on the rerun.

\subsection{Limitations of this analysis}

\begin{enumerate}
\item \textbf{Same-family atomizer.} The atomizer (\opus) is the same model family as the answer model. Extraction involves judgment, in drawing claim boundaries, preserving qualifiers, and assigning citations, and same-family bias cannot be ruled out. A cross-family atomizer (e.g., \gpt) would be the natural validation.
\item \textbf{Partial human verification.} Scoring was checked against two annotators on a stratified sample of 100 claims and did not meet the agreement rule fixed in advance, before or after the correction (\cref{app:correction}); atomization has not been checked. Remaining validation steps for a replication: (i) adjudicate the annotators' disagreements and extend the check to the decomposition arm; (ii) macro-averages by question are now reported alongside the micro-averages; (iii) separate ``support by original PDF passage'' from ``support by compiled wiki page'' --- currently wiki claims may be scored against a wiki page that itself is a rewrite of the source PDF, which conflates intermediate-representation alignment with source fidelity; (iv) report strict (supported only), loose (supported + partial), and failure (unsupported + contradicted) precision alongside the four-way breakdown.
\item \textbf{Wiki citation duplication.} Wiki sources include both wiki-page filenames and canonical PDF filenames as sibling entries; the scoring used the union of \texttt{chunk\_text} fields, which can include redundant content.
\item \textbf{Coarser than ALCE.} Our four-way categorization is less granular than ALCE's~\citep{gao2023alce} citation-precision/recall decomposition. A future replication should adopt ALCE verbatim.
\end{enumerate}

\section{Preregistered exploratory analyses (full data)}
\label{app:exploratory}

The OSF preregistration listed seven secondary analyses that are reported but do not adjudicate the H1/H2/H3 decisions. \cref{sec:results} summarizes the cost-vs-quality, latency, adaptive-thinking, and judge-notes findings; the per-question score table is in \cref{app:per-question}; the IRR analysis is in \cref{app:irr}. We give the remaining tables here.

\paragraph{Bias-tag stratification (manipulation-design check).}

\begin{table}[h]
\centering
\small
\begin{tabular}{lrrrrr}
\toprule
Bias & $n$ & $\Delta$\textsubscript{ground} & $\Delta$\textsubscript{struct} & $\Delta$\textsubscript{conflict} & $\Delta$\textsubscript{mapping} \\
\midrule
\texttt{rag} (= bias-check tier) & 3 & $-2.00$ & $+1.67$ & $+0.00$ & $+0.00$ \\
\texttt{wiki} (= synthesis tiers) & 8 & $-0.25$ & $+2.00$ & $+2.12$ & $+5.75$ \\
\texttt{neutral} (= chronological tier) & 2 & $-0.50$ & $+3.00$ & $+2.00$ & $+3.50$ \\
\bottomrule
\end{tabular}
\end{table}

The taxonomy holds: \texttt{rag}-favored questions show RAG winning \texttt{groundedness} by $2.0$; \texttt{wiki}-favored questions show Wiki ahead on three criteria and slightly behind on \texttt{groundedness} ($-0.25$).

\paragraph{Latency, thinking engagement, and cost-vs-quality clusters.}

\begin{table}[h]
\centering
\small
\begin{tabular}{lrrr}
\toprule
Metric & \rag & \wiki & ratio \\
\midrule
Total wall-clock (sum) & 3.3 min & 22.0 min & $6.6\times$ \\
Mean per-question latency & 15.4 s & 101.3 s & $6.6\times$ \\
Questions with thinking\_tokens $> 0$ & $2/13$ & $13/13$ & --- \\
Total thinking tokens & 402 & 9{,}373 & $23\times$ \\
Per-question total-token range & $5{,}019$--$7{,}721$ & $29{,}623$--$213{,}460$ & --- \\
Per-question rubric-mean range & $3.75$--$8.25$ & $6.75$--$9.25$ & --- \\
\bottomrule
\end{tabular}
\end{table}

The wall-clock ratio is smaller than the token ratio. Both systems use the same \texttt{xhigh} flag, so the flag does not account for the difference in thinking engagement; what does was not measured. Cost-vs-quality clusters are not globally dominated: 4 of 13 questions show RAG matching or beating Wiki on rubric mean at an order of magnitude fewer tokens (T1-cardio, T3-rwd, B2, B3).

\paragraph{Judge notes thematic review.}
Four recurring themes across 13 primary-judge \texttt{notes} fields:
(1) style preference for unified prose --- wiki narrative answers score higher on \texttt{structural\_integrity} than RAG bullet-and-cite style even with errors;
(2) wiki paraphrase introduces drift --- six notes mention wiki ``paraphrasing'' or ``losing precision'' on multi-hop / bias-check tiers;
(3) wiki occasionally corrects the question (T1-cardio-baselines: wiki disambiguated DEVOTE 3 trial generations; \gpt rewarded, \gem penalised; drives $\sim 1$ point of T1 IRR delta);
(4) thinking-budget asymmetry visible to judges (B3-mia, T4-cross-domain: wiki's tool-loop format reads as ``more deliberate reasoning'').

{\small\section{Preregistration: relevant excerpts}
\label{app:prereg}

The complete OSF preregistration (15 pages) is the registration record on the OSF project. The three hypotheses are quoted in \cref{sec:methods}. Repository tag: \texttt{osf-prereg-2026-05-06} (commit \texttt{655ef12}).

\paragraph{Decision quantities (verbatim).}
\begin{align*}
\text{mean\_tier}[s, c, t] &= \frac{1}{n_t} \sum_{q \in \text{tier } t} \text{score}[s, c, q] \\
\Delta_{\Hone}[c] &= \text{mean\_tier}[\text{wiki}, c, \text{multi-hop+emergence}] - \text{mean\_tier}[\text{rag}, c, \text{multi-hop+emergence}] \\
\Delta_{\Htwo} &= \text{mean\_tier}[\text{rag}, \texttt{groundedness}, \text{bias-check}] - \text{mean\_tier}[\text{wiki}, \texttt{groundedness}, \text{bias-check}]
\end{align*}

\paragraph{IRR adjustment rule.}
\emph{If $\text{max\_delta}[c] > 2$ for any criterion $c$, that criterion's \texttt{mean\_tier} values are recomputed using the mean of the two judges' scores (averaged per-question first, then aggregated by tier) in place of the primary-judge-only mean for the H1/H2 hypothesis tests; the unadjusted means are reported alongside.}

\paragraph{Robustness-check thresholds.} The preregistered Bayesian ``strongly corroborated'' threshold is $P(\mu \ge \theta) \ge 0.95$ ($\theta = +2$ for \Hone criteria, $\theta = 0$ for \Htwo). A Bonferroni-corrected sidecar at $\alpha = 0.0167$ per hypothesis is preregistered as available on reviewer request. Tail directions are explicitly one-sided; directionally opposite outcomes are reported as strong negative results.

\paragraph{Foreknowledge.} The preregistration discloses partial RAG-side ingestion (10 of 24 papers) and two pre-formal smoke tests (a single ad-hoc query against one paper; a judge smoke against synthetic Q\&A fixtures). Neither contributed analysis-relevant data.
}
{\let\scriptsize\tiny\sloppy\section{Reproducibility}
\label{app:reproducibility}

\begin{table}[h]
\centering
\small
\begin{tabular}{p{9.5cm}p{4.5cm}}
\toprule
Item & Status \\
\midrule
Code released & \checkmark{} (\coderelease) \\
Data released & \checkmark{} (artifact link; the corpus PDFs are not redistributed) \\
Run artifacts deposited with checksums & \checkmark{} (5 preregistered, 8 post-hoc) \\
Reproducer scripts that recompute every reported number & \checkmark{} (listed below) \\
Preregistration with timestamp & \checkmark{} (tag \texttt{osf-prereg-2026-05-06}) \\
Random seeds documented & \checkmark{} (NUTS and bootstrap 42; blinding 42/43; human-check sample 20260919) \\
Compute requirements stated & \checkmark{} (below) \\
Model versions stated & \checkmark{} (\opus, \gpt, \gem; exact API model IDs in the artifacts) \\
Manipulation check passed & \checkmark{} (zero exclusions) \\
Bayesian convergence diagnostics & \checkmark{} (\cref{app:bayes}) \\
Limitations enumerated & \checkmark{} (\cref{sec:limitations}) \\
Human evaluation & partial (claim-level scorer only; \cref{app:correction}) \\
Second scorer for the claim-level analysis & $\times$ (single scorer) \\
\bottomrule
\end{tabular}
\end{table}

\paragraph{Compute requirements.}
\begin{itemize}
\item RAG ingestion: $\sim$2 h wall-clock on a single machine with API access; $\sim$\$33 USD in API spend.
\item RAG run (13 queries): $\sim$3 minutes wall-clock; $\sim$\$2 USD.
\item Wiki ingestion: $\sim$7 h active wall-clock (per-paper stopwatch); cost recorded externally.
\item Wiki run (13 queries): $\sim$22 minutes wall-clock; cost recorded in the wiki-run telemetry.
\item Judge run (primary + IRR over 13 questions): $\sim$5 minutes wall-clock; $\sim$\$1.50 USD combined.
\item Preregistered analysis (NUTS sampling): $\sim$2 minutes on a 4-core CPU.
\item Post-hoc claim-level grounding (atomize + score 466 claims): $\sim$30 minutes wall-clock; $\sim$\$2 USD.
\item Claim-level rescoring on full passages: 505 scorer calls for the main arms and their capped control (325{,}103 input and 112{,}754 output tokens), and 219 for decomp-RAG (196{,}898 and 53{,}134).
\end{itemize}

\paragraph{Artifact integrity.} SHA-256 hashes for every deposited artifact, as listed in \texttt{ARTIFACTS\_MANIFEST.txt} in the supplement and at the artifact link. Verify with \texttt{sha256sum <filename>}.

\begin{table}[h]
\centering
\scriptsize
\begin{tabular}{ll}
\toprule
File & sha256 \\
\midrule
\texttt{ingest-20260506T202904Z.json} & \texttt{4a5ff508dc449ecb934a40e643ed524d125e383ef10000564dfbf9b84db73c28} \\
\texttt{run-20260506T221602Z.json} & \texttt{eb295306f91e39b764a02d64c8b745591087aec9d5f5d727ef132ba28b98537d} \\
\texttt{wiki-ingest-20260506T211731Z.json} & \texttt{da22f3aded2b747ae360f9eda91d8a64bacd43406f7bdfc3fee144ff73b86657} \\
\texttt{wiki-run-20260506T205500Z.json} & \texttt{6c1da6be7f7d5655b24152b3636be81b2835a94a35f593572f86211c4294ce8b} \\
\texttt{judged-20260506T225645Z.json} & \texttt{b74f44a75252ec3696e07ac47bcb6edc93ba90187fe9db70763694766deab163} \\
\texttt{post-hoc/grounding-20260508T155234Z.json} & \texttt{33c2da6447e54c16cfba62c2621f507f22b39e67fdd81cfb4fb7b21c40a88451} \\
\texttt{post-hoc/run-decomp-20260509T001618Z.json} & \texttt{240ddcb870fcbbfba99082b7b3b76078913f9dea779b4cc7897edacbd66a5884} \\
\texttt{post-hoc/judged-20260509T012315Z.json} & \texttt{ae252310f8cb2f1150c7b58e5273ad2235ad48d558ab4ec496e776a4b49ff976} \\
\texttt{post-hoc/grounding-decomp-20260509T021528Z.json} & \texttt{986257981da1f8d22bc2de7bb75493f0b4c145157a9a5ea3cc2f5254dade6994} \\
\texttt{post-hoc/judged-20260731T145648Z.json} & \texttt{3330a3e2610e4f19da4c836056b00f4d5de0fc9749687b081df8e1fb31d9c616} \\
\texttt{post-hoc/rescore-main-capnone-20260928T113149Z.json} & \texttt{ec0e1937c47f88e6e7c88db1e75acdb3793cdcf50bf09517d09eac38d813ee61} \\
\texttt{post-hoc/rescore-main-cap1500-rag-20260928T113351Z.json} & \texttt{02b42987cfe073d23ed82631a30ffed7832c60b15e2930da2b936364aa301dc4} \\
\texttt{post-hoc/rescore-decomp-capnone-20260928T201439Z.json} & \texttt{8b04b4676003f82b33385f2261cd2bae3b53f94cd75b89a986c3b57895cca4e6} \\
\bottomrule
\end{tabular}
\end{table}

\paragraph{Reproducing every number in the paper.} The supplementary archive is self-contained: its \texttt{experiments/results/} holds the deposited artifacts, and every script below runs from the archive root.
\begin{verbatim}
pip install pymc numpy

# Preregistered analyses and robustness checks (Sections 4 and 5)
PYTENSOR_FLAGS="cxx=" python experiments/run_preregistered_analysis.py
python experiments/analyze_irr_stats.py      # agreement statistics
python experiments/reconstruct_h3a.py        # H3a billing sensitivity

# Claim-level figures, first-reported and corrected (5.4, 6, App. H)
python experiments/claim_level_figures.py \
  --main experiments/results/post-hoc/rescore-main-capnone-20260928T113149Z.json \
  --decomp experiments/results/post-hoc/rescore-decomp-capnone-20260928T201439Z.json

# Human check of the claim-level scorer (App. H)
# follow experiments/human_validation/README.md
\end{verbatim}
Re-running any model call (\texttt{run\_claim\_grounding.py}, \texttt{rescore\_claim\_grounding.py}, \texttt{run\_decomp\_rag.py}) requires API keys and produces verdicts that differ from run to run; the capped rerun in \cref{app:correction} measures that variation at 10\% to 14\% of claim-level verdicts.
}
\section{Correction to the claim-level analysis, and a human check of its scorer}
\label{app:correction}

This appendix documents the error in the post-hoc claim-level analysis of \cref{sec:results-claim}, the rescoring that corrects it, and the human check that exposed it. The human check was designed, and its decision rule fixed in writing, before any label existed. Everything else here is post-hoc.

\subsection{The truncation}

The scoring script, \texttt{experiments/run\_claim\_grounding.py}, cut every cited passage to its first 1{,}500 characters before passing it to the scorer, and the paper as first reported did not disclose the cut. Cited RAG passages are retrieved chunks of about 2{,}500 characters and cited wiki passages are page excerpts of about 430, so the cut reached 128 of the 132 cited single-round RAG claims, 200 of the 219 cited decomp-RAG claims, and none of the 241 cited wiki claims. RAG claims were therefore scored against partial evidence and wiki claims against complete evidence. The rescoring passes the full concatenated passages with the original scorer prompt, the original atomized claims, and the original scorer model, \gpt at \texttt{medium} reasoning, served as \texttt{gpt-5.4-2026-03-05}; the first run recorded only the model alias. The atomization step saw each source cut to 600 characters. It takes citations from the answer's own attributions, and it was not rerun.

\subsection{Separating the cut from rerun noise}

\begin{table}[h]
\caption{Cited-claim rates as first reported, on a capped rerun, and on full passages. The wiki's passages were never cut, so its rerun is also its full-passage rescoring. No capped control was run for decomp-RAG.}
\label{tab:rerun-control}
\centering
\small
\begin{tabular}{lrrr}
\toprule
 & Verdicts changed & Supported (\%) & Unsupported (\%) \\
\midrule
RAG, first run (cut) & --- & 18.9 & 34.1 \\
RAG, rerun with the cut & 19 of 132 & 22.7 & 36.4 \\
RAG, rerun on full passages & 57 of 132 & 53.8 & 12.1 \\
\addlinespace
Decomp-RAG, first run (cut) & --- & 19.2 & 41.6 \\
Decomp-RAG, rerun on full passages & 46 of 219 & 32.4 & 32.0 \\
\addlinespace
Wiki, first run & --- & 40.2 & 6.2 \\
Wiki, rerun, inputs unchanged & 24 of 241 & 41.5 & 7.1 \\
\bottomrule
\end{tabular}
\end{table}

A plain rerun changes 10\% to 14\% of verdicts and moves the rates by up to four points. Removing the cut changes 43\% of single-round RAG verdicts and nearly triples its supported rate, which reverses the direction of the supported-rate comparison with the wiki.

\subsection{Figures as first reported and as corrected}

\begin{table}[h]
\caption{Every claim-level figure the paper prints, as first reported (from truncated evidence) and as corrected. Cited-claim and macro rates are conditional on a citation; the all-claims view counts uncited claims as unsupported. The recomputation first reproduced each first-reported figure exactly from the original verdicts.}
\label{tab:claim-correction}
\centering
\small
\begin{tabular}{lrrrrrr}
\toprule
 & \multicolumn{2}{c}{RAG} & \multicolumn{2}{c}{Decomp-RAG} & \multicolumn{2}{c}{Wiki} \\
\cmidrule(lr){2-3}\cmidrule(lr){4-5}\cmidrule(lr){6-7}
 & first & corrected & first & corrected & first & corrected \\
\midrule
Cited claims supported (\%) & 18.9 & 53.8 & 19.2 & 32.4 & 40.2 & 41.5 \\
Cited claims unsupported (\%) & 34.1 & 12.1 & 41.6 & 32.0 & 6.2 & 7.1 \\
Macro by question, supported (\%) & 18.4 & 51.1 & 17.0 & 28.8 & 37.3 & 37.9 \\
Macro by question, unsupported (\%) & 35.8 & 12.1 & 50.6 & 41.7 & 8.0 & 8.7 \\
All claims, supported (\%) & 16.7 & 47.3 & 18.1 & 30.6 & 30.7 & 31.6 \\
All claims, unsupported (\%) & 42.0 & 22.7 & 44.8 & 35.8 & 28.5 & 29.1 \\
Contradicted (count) & 3 & 4 & 2 & 1 & 1 & 1 \\
\bottomrule
\end{tabular}
\end{table}

\begin{table}[h]
\caption{Corrected claim-level rates by tier, conditional on a citation, with the first-reported supported rate for comparison. Decomp-RAG counts a claim as cited when its citation resolves to a chunk, as in \cref{sec:ablation}.}
\label{tab:claim-tiers}
\centering
\small
\begin{tabular}{llrrrrrr}
\toprule
Tier & System & $n$ & Sup.\ first (\%) & Sup.\ (\%) & Part.\ (\%) & Unsup.\ (\%) & Contr.\ (\%) \\
\midrule
bias-check & RAG & 19 & 5.3 & 36.8 & 21.1 & 36.8 & 5.3 \\
 & Decomp & 23 & 43.5 & 60.9 & 26.1 & 13.0 & 0.0 \\
 & Wiki & 27 & 51.9 & 51.9 & 40.7 & 7.4 & 0.0 \\
\addlinespace
chronological & RAG & 25 & 8.0 & 56.0 & 24.0 & 20.0 & 0.0 \\
 & Decomp & 24 & 8.3 & 20.8 & 41.7 & 37.5 & 0.0 \\
 & Wiki & 61 & 50.8 & 50.8 & 45.9 & 3.3 & 0.0 \\
\addlinespace
conflict & RAG & 22 & 22.7 & 63.6 & 31.8 & 0.0 & 4.5 \\
 & Decomp & 50 & 14.0 & 28.0 & 28.0 & 42.0 & 2.0 \\
 & Wiki & 35 & 42.9 & 45.7 & 42.9 & 11.4 & 0.0 \\
\addlinespace
emergence & RAG & 20 & 25.0 & 50.0 & 25.0 & 20.0 & 5.0 \\
 & Decomp & 37 & 8.1 & 29.7 & 43.2 & 27.0 & 0.0 \\
 & Wiki & 32 & 37.5 & 31.2 & 65.6 & 0.0 & 3.1 \\
\addlinespace
multi-hop & RAG & 25 & 32.0 & 76.0 & 24.0 & 0.0 & 0.0 \\
 & Decomp & 46 & 37.0 & 39.1 & 30.4 & 30.4 & 0.0 \\
 & Wiki & 45 & 35.6 & 40.0 & 48.9 & 11.1 & 0.0 \\
\addlinespace
policy & RAG & 21 & 19.0 & 33.3 & 61.9 & 0.0 & 4.8 \\
 & Decomp & 39 & 7.7 & 23.1 & 43.6 & 33.3 & 0.0 \\
 & Wiki & 41 & 22.0 & 26.8 & 63.4 & 9.8 & 0.0 \\
\bottomrule
\end{tabular}
\end{table}

\subsection{The human check}

\paragraph{Design, fixed in advance.} One hundred of the 373 cited single-round RAG and wiki claims, 50 per system, stratified by question tier and drawn at a fixed seed. Two annotators who are not authors labeled them independently. Each saw the claim, the full passages it cited, and the scorer's verdict definitions (\cref{app:claimgrounding}), including the instruction to be strict on the supported-partial boundary. Neither was told the research question, that two systems were compared, or which system produced an item. The decision rule: the scorer counts as validated only if its binary agreement with the annotators, supported against not supported, reaches 70\% and Cohen's $\kappa$ reaches 0.40.

\paragraph{Agreement.} The annotators agree with each other on 94 of the 100 claims on the binary cut ($\kappa = 0.81$, 95\% CI for agreement 90\% to 98\%) and on 91 across all four verdicts ($\kappa = 0.74$). The scorer as first run agrees with them on 54 and 50 claims ($\kappa = 0.23$ and $0.19$), and on 52 and 47 across four verdicts. The corrected scorer agrees on 69 and 65 ($\kappa = 0.40$ and $0.33$). Against each annotator separately, the scorer does not meet the rule before or after the correction. Agreement with a consensus is taken up under Adjudication below.

\paragraph{Why the scorer disagrees.} Two causes are measured; their ranking is judgment. The first is the truncation: it cut 49 of the 50 sampled RAG items and none of the 50 wiki items, and on RAG items the first-run scorer agrees with the annotators on only 17 of 50. Fourteen RAG items were scored unsupported by the first run and supported by both annotators; all 14 had a passage longer than the cut, and in 13 of them some of the claim's key terms appear only after character 1{,}500. The corrected scorer calls 10 of these 14 supported, 4 partial, and none unsupported. The second cause is a stricter reading of the supported-partial boundary in both systems: the first-run scorer used \texttt{partial} for 40 of the 100 claims, where the annotators used it 10 and 3 times. This cause does not depend on the cut. After the correction, every remaining binary disagreement is a claim the annotators call supported and the scorer calls partial or, in a few cases, contradicted.

\begin{table}[h]
\caption{Cited-claim rates on each annotator's labels, $n=50$ per system. $p_\text{conv}$: two-sided pooled two-proportion test. $p_\text{fpc}$: the test fixed in the analysis script on 2026-09-19, before any label existed, a two-sided $z$ test using unpooled variance estimates, each with a finite-population correction for the 132 RAG and 241 wiki cited claims sampled. The correction supports inference about these claims rather than about the systems, and the pre-specification names it only in its power calculation. Both treat claims as independent although claims cluster within answers, so both overstate precision.}
\label{tab:human-rates}
\centering
\small
\begin{tabular}{lrrrr}
\toprule
 & RAG & Wiki & $p_\text{conv}$ & $p_\text{fpc}$ \\
\midrule
Supported, annotator 1 & 78\% & 78\% & 1.0 & 1.0 \\
Supported, annotator 2 & 78\% & 86\% & 0.30 & 0.21 \\
Unsupported, annotator 1 & 16\% & 6\% & 0.11 & 0.049 \\
Unsupported, annotator 2 & 16\% & 10\% & 0.37 & 0.28 \\
\bottomrule
\end{tabular}
\end{table}

\paragraph{Adjudication was not done.} The design called for adjudicating the annotators' disagreements, 6 on the binary cut and 9 across four verdicts, into a consensus. We did not produce it, and each annotator is reported separately. Adjudication cannot change the verdict on the first-run scorer, since any consensus agrees with it on between 49 and 55 of the 100 claims. It could change the verdict on the corrected scorer. That scorer agrees with both annotators on 64 claims and with exactly one of them on each of the 6 disputed claims, so a consensus would agree with it on between 64 and 70, and the upper end reaches the 70\% floor. The corrected scorer therefore fails the rule against each annotator separately, and its validation against a consensus remains unassessed. Adjudication could also change the unsupported-rate contrast, whose conventional $p$-value would be 0.11, 0.22, or 0.37 depending on how two disputed wiki items were ruled (0.049, 0.13, or 0.28 by the finite-population test).

\paragraph{Limits.} The check is not fully blind: RAG passages are longer, and a single length threshold separates 97\% of the items by system, although the annotators were not told that two systems existed. Like the scorer, it measures whether a claim follows from the evidence the system cited, not whether it is faithful to the original paper. It evaluates the support label for each claim as extracted; it does not check the extraction or the citation assignment, which ran with each source cut to 600 characters. It covers the claim-level scorer on single-round RAG and wiki claims only, not the rubric judges or the decomposition arm.

\paragraph{Materials.} The three rescoring outputs are deposited at the artifact link, bringing the deposit to 13 artifacts. The pre-specification, the sampling and analysis scripts, the rescoring script, and both annotators' labels are in the supplementary material.

\section{Decomposition prompt and tier construction}
\label{app:decomp-tiers}

\subsection{Decomposition prompt}

The decomposer is \opus at \texttt{medium} reasoning effort, called once per question with the question text as the user message and the system prompt below, verbatim. The code checks only that the reply is a JSON list of two to five strings; no separate step validated the content of the sub-questions.

\begin{small}
\begin{verbatim}
You are a careful research analyst. Given a research question, decompose it into
2 to 5 sub-questions whose answers, taken together, would let a well-informed reader
answer the original question.

Rules:
- Each sub-question must be self-contained and answerable from a research-paper
corpus (no pronoun references back to the original).
- Sub-questions should not overlap heavily; each should target a distinct piece of
evidence the original needs.
- If the original is a simple point-source lookup ("what was the confidence interval
in DEVOTE 3?"), still produce 2-3 sub-questions that target adjacent context the
answer should cite.
- Do NOT answer the original question. Only emit the sub-question list.

Output ONLY a JSON array of strings: ["sub-q 1", "sub-q 2", ...]. No prose, no markdown.
\end{verbatim}
\end{small}

\subsection{Tier construction}

The question set was built to probe the wiki's hypothesized strengths, and a tier of point-source questions expected to favor RAG was included to probe point-source retrieval alongside them. With eight questions labeled wiki-favoring, three RAG-favoring, and two neutral, the set is neither balanced nor representative, and the tiers are task categories rather than an ordered scale of difficulty. The tiers, as the frozen question file names them, are chronological evolution (whether a system updates earlier findings with later ones rather than appending them), conflict detection, multi-hop technical synthesis, emergence or discovery across the corpus, policy and strategy, and bias-check point-source retrieval. Each question carries an expected-bias label recorded in the same file before any run.

\begin{table}[h]
\caption{The 13 questions by tier, with the expected-bias label recorded in the question file frozen at the preregistration tag.}
\label{tab:tiers}
\centering
\small
\begin{tabular}{lll}
\toprule
Question & Tier & Expected bias \\
\midrule
T1-cardio-baselines & chronological & neutral \\
T1-mia-evolution & chronological & neutral \\
T2-oae-bio-vs-chemical & conflict & wiki \\
T2-fair-use-vs-optin & conflict & wiki \\
T3-mia-as-copyright-evidence & multi-hop & wiki \\
T3-rwd-validity-for-side-effects & multi-hop & wiki \\
T4-eelgrass-mrv-gaps & emergence & wiki \\
T4-cross-domain-audit & emergence & wiki \\
T5-glp1-policy & policy & wiki \\
T5-25th-paper & policy & wiki \\
B1-devote3-confidence-interval & bias-check & RAG \\
B2-he-tyka-equilibration-ratios & bias-check & RAG \\
B3-mia-attack-success-rate & bias-check & RAG \\
\bottomrule
\end{tabular}
\end{table}

We do not claim the tiers are exhaustive, or that each question isolates the property its tier names. With two or three questions per tier, tier-level results in this paper are descriptive.

\end{document}